\documentclass[preprints,article,accept,oneauthor,pdftex]{mdpi} 

\firstpage{1} 
\pubvolume{1}
\issuenum{1}
\articlenumber{0}
\pubyear{2021}
\copyrightyear{2020}
\datereceived{} 
\dateaccepted{} 
\datepublished{} 
\hreflink{https://doi.org/} 

\usepackage[utf8]{inputenc} 
\usepackage[T1]{fontenc}    
\usepackage{url}            
\usepackage{booktabs}       
\usepackage{amsfonts}       
\usepackage{nicefrac}       
\usepackage{microtype}      

\usepackage{amsmath,amsthm}
\usepackage{amssymb,graphicx}
\usepackage{algorithm}
\usepackage{algorithmic}

\usepackage[capitalize]{cleveref}
\crefname{section}{Sec.}{Secs.}
\Crefname{section}{Section}{Sections}
\Crefname{table}{Table}{Tables}
\crefname{table}{Tab.}{Tabs.}

\graphicspath{{figs/}{pics/} {pics/filters/}}
\DeclareGraphicsExtensions{.jpg,.pdf,.mps,.png}
\def\x{{\mathbf x}}

\def\RR{\mathbb R}

\newcommand{\ba}{\mathbf{a}}
\newcommand{\bA}{\mathbf{A}}
\newcommand{\bb}{\mathbf{b}}
\newcommand{\bB}{\mathbf{B}}
\newcommand{\bc}{\mathbf{c}}
\newcommand{\bC}{\mathbf{C}}
\newcommand{\bD}{\mathbf{D}}

\newcommand{\bF}{\mathbf{F}}

\newcommand{\bG}{\mathbf{G}}

\newcommand{\bI}{\mathbf{I}}

\newcommand{\bM}{\mathbf{M}}
\newcommand{\bo}{\mathbf{1}}

\newcommand{\bS}{\mathbf{S}}
\newcommand{\bx}{\mathbf{x}}
\newcommand{\bX}{\mathbf{X}}
\newcommand{\by}{\mathbf{y}}
\newcommand{\bY}{\mathbf{Y}}

\newcommand{\bU}{\mathbf{U}}
\newcommand{\bv}{\mathbf{v}}
\newcommand{\bV}{\mathbf{V}}

\Title{Training a Two Layer ReLU Network Analytically}
\TitleCitation{Training a Two Layer ReLU Network Analytically}

\Author{Adrian Barbu $^{1}$\orcidA{}*}

\AuthorNames{Adrian Barbu}

\AuthorCitation{Barbu, A.}

\address{%
$^{1}$ \quad Statistics Department, Florida State University, Tallahassee, FL 32306, USA; abarbu@fsu.edu}

\corres{Corresponding author}
\abstract{
Neural networks are usually trained with different variants of gradient descent based optimization algorithms such as stochastic gradient descent or the Adam optimizer.
Recent theoretical work states that the critical points (where the gradient of the loss  is zero) of two-layer ReLU networks with the square loss are not all local minima.
However, in this work we will explore an algorithm for training two-layer neural networks with ReLU-like activation and the square loss that alternatively finds the critical points  of the loss function analytically for one layer while keeping the other layer and the neuron activation pattern fixed. 
Experiments indicate that this simple algorithm can find deeper optima than Stochastic Gradient Descent or the Adam optimizer, obtaining significantly smaller training loss values on four out of the five real datasets evaluated. 
Moreover, the method is faster than the gradient descent methods and has virtually no tuning parameters.
}


\begin{document}

\section{Introduction}

\label{sec:intro}

Training neural networks is usually done using different gradient descent-based algorithms such as stochastic gradient descent or the Adam optimizer  \cite{kingma2014adam}. 
This type of training involves many passes through the entire data, usually on the order of 100-300, which makes it very slow. 
Moreover, the results of this training are sensitive to a number of tuning parameters such as the learning rate and the minibatch size, as well as the manner in which these parameters are changed during the training iterations (e.g. the learning rate schedule). 
For these reasons, many training runs are usually conducted (on the order of 10 or more) with different parameter and schedule combinations and the most successful one is used to obtain the final model. 
The fact that the training takes at least 100 epochs and usually at least 10 training runs are used to find a good combination, it implies that to train a neural network well, one must use at least 1000 passes through the data, which can be computationally expensive.

These computational reasons serve as motivation for studying a novel method for training two layer neural networks with the square loss that needs as little as 10 passes through all the data. 
The method is capable of obtaining a loss that is smaller than the losses obtained by the well-tuned gradient descent algorithms and has better generalization when the number of inputs is small.

In \cite{sun2020global} is mentioned that ``practitioners find that narrow neural nets cannot be solved well''. 
Our works confirms this statement experimentally at least for NNs with a small number of inputs and shows that the proposed approach can find better local minima of the loss.
This work could enable a wider use of shallow NNs for many computation restricted applications.
The authors of \cite{sun2020global} also point out that the test error of an algorithm can be decomposed into the representation error, optimization error and generalization error. 
This work  addresses the optimization error for a narrow type of NNs, namely the two layer NNs with ReLU-like activation and the square loss.
Even though the scope is narrow, we show that in some cases the optimization capabilities of the proposed method greatly outperform those of the standard gradient descent methods.

\begin{figure}[t]
\centering
\includegraphics[width=0.4\linewidth]{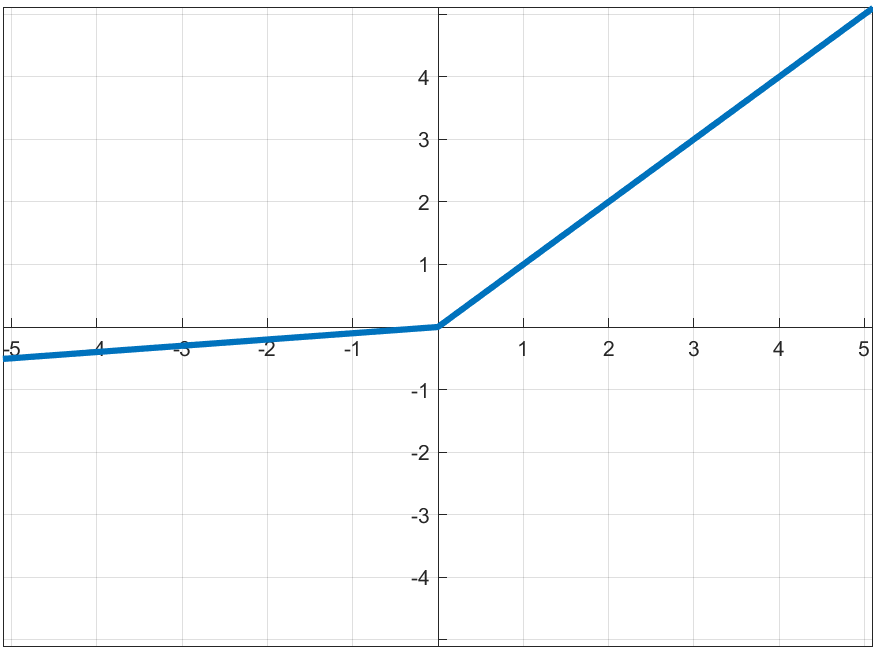}
\includegraphics[width=0.4\linewidth]{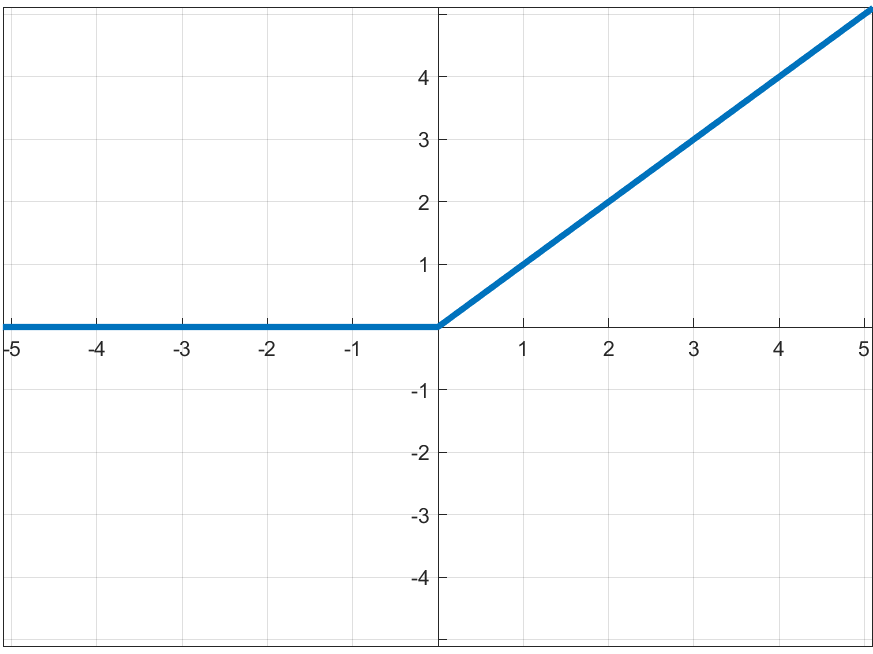}
\vspace{-1mm}
\caption{This work focuses on 2-layer NNs with leaky ReLU-like activation functions $\sigma(x)=\alpha x+ (1-\alpha)\max(0,x)$, with $\alpha\in [0,1)$. Shown are the leaky ReLU ($\alpha=0.1$, left) and the ReLU ($\alpha=0$, right). }\label{fig:relu}
\end{figure}

The contributions of this work are the following:
\begin{itemize}
\item It introduces a method for training a two layer NN with leaky ReLU-like activation and the square loss analytically, by solving ordinary least squares (OLS) equations alternatively for each layer while keeping the activation pattern of the neurons fixed. 
The neuron activation pattern is a binary matrix indicating what hidden neurons fire (have non-zero activation) for each observation. 
It is described in Section \ref{sec:fitA}.
\item It conducts experiments on real and simulated data to see in what conditions the proposed method outperforms standard gradient-based optimization such as the standard stochastic gradient descent, the Adam optimizer and the LBFGS algorithm. 
These experiments indicate that the proposed analytic method is faster and obtains lower loss minima when the number of observations is not very large or when the input dimension is small.
\end{itemize}

The paper is organized as follows: Section \ref{sec:relwork} gives an overview of the related work. 
Section \ref{sec:methods} introduces the proposed analytic minimization method that involves fitting each layer while keeping the other layer fixed. 
The experimental results on real and simulated data, as well as an ablation study, are presented in Section \ref{sec:results}. 
The paper finishes with conclusions and future work in Section \ref{sec:conclusion}.

\subsection{Related Work} \label{sec:relwork}

Critical points of deep linear NNs (with no activation function) with the square loss have been studied in \cite{zhou2018critical}. 
For deep linear NNs, the authors proved that every local minimum is also a global minimum and all other critical points are saddle points.
The authors have also studied the critical points of one hidden layer NNs with ReLU activation and the square loss, but only in certain regions of the parameter space. 
They have characterized the critical points in the entire space only for a NN with one hidden node. 
Moreover, they have made no attempt to provide an explicit algorithm for finding the critical points. 
In contrast, this work presents a simple and efficient algorithm that finds the critical points for the two layers alternatively while keeping the neuron firing pattern fixed.

Recent work \cite{gopalani2022global} has shown that training certain two-layer NNs with the square loss and Stochastic Gradient Descent and adequate regularization can find the global optimum of the loss function. 
However, these networks must have smooth and bounded activation functions such as the sigmoid or tanh activation and not the ReLU activation. 
The proof uses a Poincare type inequality for Villani functions \cite{villani2006hypocoercivity,shi2020learning}, which was proved in \cite{shi2020learning}.

Some works \cite{zhang2017convexified,pilanci2020neural} relax the NNs to be able to use convex optimization and obtain global solutions. 
The first work \cite{zhang2017convexified} is aimed at CNNs where the shared filters are represented as low rank constraints, which are relaxed for convex optimization. 
The second work \cite{pilanci2020neural} focuses on two layer ReLU networks and using convex duality it shows that the two layer ReLU network can be globally optimized with a second order cone program with a computation complexity of $O\left (d^3r^3\left (\frac{N}{r}\right )^{3r}\right)$ where $r$ is the rank of the input data matrix (which is usually $r=d$ for large $N$). 
This is polynomial time in sample size $N$ but exponential in the input dimension $d$, so it can be prohibitive for large $N$ and $d$. 
They also prove that the exponential complexity in $d$ cannot be improved unless P=NP. 
In contrast, the algorithm introduced in this paper is faster, being $O((dh)^3+N(dh)^2)$, where $h$ is the number of hidden nodes, but it does not guarantee a global optimum.
The convex relaxation idea was further improved in \cite{mishkin2022fast}, where faster approximate algorithms were proposed based on gated ReLU models. 
However the algorithm is gradient based, relying on the accelerated proximal gradient method.
To their advantage, the method can be applied to arbitrary loss functions, not only the square loss needed for our method.

\begin{table}[t]
\centering
\begin{tabular}{l|ccc|c|c}
 &\multicolumn{3}{|c|}{Complexity in} &Loss&Global optimum \\
Method &$N$ &$d$ &$h$ &fn.&guarantee \\
\hline
Adam \cite{kingma2014adam} &$N$ &$d$ &$h$  &any&No\\
SGD \cite{gopalani2022global} &$N$ &$d$ &$h$  &$\ell_2$&Yes (sigmoid and tanh activation)\\
ICA based \cite{bakshi2019learning,goel2019learning}  &poly($N$) &poly($d$) &$\exp(h)$&some &No\\
Convex duality \cite{pilanci2020neural} &$N^d$ &$\exp(d)$ &- &$\ell_2$ &Yes\\
TD based \cite{awasthi2021efficient} &poly($N$) &poly($d$) &poly($h$) &- &No\\
Proposed &$N$ &$d^3$ &$h^3$ &$\ell_2$ &No\\
\hline
\end{tabular}
\caption{Overview of related works with time complexity and theoretical global optimum guarantees.}\label{tab:works}
\end{table}

Other works \cite{bakshi2019learning,goel2019learning} try to develop algorithms for learning NNs in polynomial time under the assumption that the bias terms of the hidden nodes are 0. 
In \cite{bakshi2019learning} the authors focus on two layer NNs with ReLU activation under the assumption that the true weight matrix of the hidden nodes is full rank. 
They give an algorithm that is polynomial in $N$ and $d$ but exponential in the number of hidden nodes $h$ that guarantees to find the true coefficients under certain assumptions. 
However, the algorithm is based on Independent Component Analysis and not on loss minimization.
Again, our algorithm is cubic in the number of hidden nodes $h$, at the cost of a lack of a global optimum guarantee.

The zero bias assumption is relaxed in \cite{awasthi2021efficient}, under the assumption that the weight matrix of the hidden nodes has linearly independent columns. 
Their algorithm is based on tensor decomposition and not on optimizing a loss function. 
It runs in polynomial time in $N$, $h$ and $d$ and recovers the true coefficients, up to a permutation and within a given error $\epsilon$, given sufficiently many samples. 
However, the most important drawback is that the algorithm assumes there is no noise in the target data, so it cannot be used in practice.
This kind of no noise assumption is also used in \cite{rolnick2020reverse}, where the authors show that it is possible to recover a deep ReLU NN including its architecture, weights and biases from the network output alone, up to an isomorphism. 
For that, they rely on the piecewise linear regions of the network output, which are defined by the neuron firing patterns that are also used in our work.

An argument could be made that the BFGS \cite{fletcher2013practical} and LBFGS \cite{liu1989limited} algorithms, which use an approximation of the inverse Hessian matrix, are somehow close competitors to the proposed method. 
However, these algorithms don't freeze the neuron firing pattern at each iteration and instead try to search for the optimum in the energy landscape of the original parameter space. 
The problem with these approaches is that the energy landscape has many points where the inverse Hessian is infinite, because of the non-differentiable nature of the piecewise linear ReLU activation.
For this reason, the BFGS and LBFGS algorithms explode near the local optima, as it can be easily observed experimentally. 
In contrast, by freezing the neuron firing pattern in our work, the analytic solution was observed experimentally to always exists for some data (three of the five real datasets from experiments) and usually exists for the rest of the data.

Table \ref{tab:works} presents an overview of the above mentioned works, with their computation complexities, types of loss function they optimize, and whether they provide global optimum guarantees or not.

\section{Materials and Methods} \label{sec:methods}

This work will focus on two layer ReLU Networks:
\[
f(\bx)=\sigma((1,\bx^T)\bA)\bB+\bb^0\in \RR^c
\]
where $\bx \in \RR^d$ is an input vector, $\bA=(\ba_1,...,\ba_h)$ is the $(d+1)\times h$ matrix of weights for the hidden layer and $\bB=(\bb_1,...,\bb_c)$ is the $h\times c$ matrix of weights of the output layer and $\bb^0=(b^0_1,...,b^0_c)$ are the biases for the $c$ outputs.
The activation functions of interest are the leaky ReLU type $\sigma(x)=\alpha x+ (1-\alpha)\max(0,x)$, with $\alpha\in [0,1)$. In experiments we will look at the ReLU ($\alpha=0$) and the leaky ReLU ($\alpha=0.1$) (Fig. \ref{fig:relu}).

Given observations $(\bx_1,\by_1),...,(\bx_N,\by_N)\in\RR^d\times \RR^c$, the square loss function for these networks is:
\begin{equation}
L(\bA,\bB, \bb^0)=\frac{1}{N}\sum_{k=1}^c\sum_{i=1}^N \|\sigma((1,\bx_i^T)\bA)\bb_k+b^0_{k}-y_{ik}\|^2
+\lambda (\|\bb^0\|_2^2+\|\bB\|_F^2+ \|\bA\|_F^2)), \label{eq:loss}
\end{equation}
where $\|A\|_F^2=\sum_{i,j} A_{ij}^2$ is the Frobenius norm.

We will use the standard notation denoting by $\bX$ the matrix of observations 
\begin{equation}
\bX=\begin{pmatrix}
1 &x_{11}&... &x_{1d}\\
&...\\
1 &x_{N1}&... &x_{Nd}
\end{pmatrix}, 
\end{equation}

and by $\bY$ the matrix of outputs, with $\by_i^T$ as rows and $\by_{\cdot j}$ as columns,
\begin{equation}
\bY=\begin{pmatrix}
\by_{1}^T\\
...\\
\by_{N}^T
\end{pmatrix}=
\begin{pmatrix}
y_{11}&... &y_{1c}\\
&...\\
y_{N1}&... &y_{Nc}
\end{pmatrix}=(\by_{\cdot 1},...,\by_{\cdot c}), 
\end{equation}
where $\by_{\cdot j}$ is the vector of targets for the $j$-th output.

We will use an algorithm that minimizes the loss \eqref{eq:loss} by alternately finding the critical points of the loss in terms of $\bB$ and $\bA$, where for the critical points w.r.t. $\bA$ the neuron firing pattern is kept frozen. The procedure is described in Algorithm \ref{alg:anmin}.
\begin{algorithm}
\caption{\bf \small{Analytic Minimization (ANMIN)}}\label{alg:anmin}
\begin{algorithmic}
\STATE {\bfseries Input:} Training data $\bX,\bY$.
\STATE {\bfseries Output:} Trained network parameter vectors $(\hat \bA,\hat \bB,\hat \bb_0)$.
\end{algorithmic}
\begin{algorithmic}[1]
\STATE Initialize $\bA$ randomly
\STATE Fit $(\bB,\bb^0)$ using eq. \eqref{eq:beta}.
\STATE Compute loss $l_0=L(\bA,\bB,\bb^0)$
\FOR {$i = 1$ to $E$}
\STATE Update $\bF=I(\bX\bA>0)$ and $\bG=(1-\alpha)\bF+\alpha \bo$
\STATE Compute $\bM$ using Eq. \eqref{eq:M} and $\bc$ using Eq. \eqref{eq:C}
\IF{$\ln |\bM+\lambda \bI_{(d+1)h}|>\tau$}
\STATE Solve $(\bM+\lambda \bI_{(d+1)h})\ba=\bc$ 
\ELSE
\STATE Obtain $\bU,\bD,\bV$ by SVD such that \\$\bU\bD\bV^T=\bM+\lambda \bI_{(d+1)h}$
\STATE If $D_{ii}<0.0001$, set $D_{ii}=0.0001, i=1,...,dh$
\STATE Obtain $\ba=\bU\bD^{-1}\bV^T\bc$ 
\ENDIF
\STATE Reshape vector $\ba$ into $(d+1)\times h$ matrix $\bA$
\STATE Fit $(\bB,\bb^0)$ using eq. \eqref{eq:beta}.
\STATE Compute loss $l_i=L(\bA,\bB,\bb^0)$
\IF {$l_i<l_0$}
\STATE Set $\hat \bA=\bA,\hat \bB=\bB,\hat \bb^0=\bb^0$
\STATE Set $l_0=l_i$
\ENDIF
\ENDFOR
\end{algorithmic}
\end{algorithm}

The threshold $\tau$ controls when the linear system $(\bM+\lambda \bI_{(d+1)h})\ba=\bC$ can be solved numerically. In practice we take $\tau=-10,000$.

\subsection{Fitting the Output Layer Weights $\bB$}

Fitting the output layer weight matrix $\bB$ is simple OLS, using $\bS=(\sigma(\bX\bA),\bo_N)$ as input. Because the square loss is a sum 
 of 
  the square losses over the output variables, $\bB$ can be solved separately for each output variable:

\begin{equation}
(\frac{1}{N}\bS^t\bS+\lambda \bI_{h+1})\begin{pmatrix}\bb_k\\b^0_k\end{pmatrix}=\bS^T\by_{\cdot k}, k=1,...,c \label{eq:beta}
\end{equation}

\subsection{Fitting the Hidden Layer Weights $\bA$} \label{sec:fitA}

Let $\bF=I(\bX\bA>0)$ be the $N\times h$ binary firing pattern matrix, indicating what hidden neurons fire for each observation. 
Here $I(x)$ is the elementwise indicator operator.
Let $\bG=(1-\alpha)\bF+\alpha \bo$, where $\bo$ is the matrix with all entries $1$.

Then the neural network response for training observation $(\bx_i,\by_i)$,  is 
\[
\hat y_{ik}\hspace{-0.5mm}=\hspace{-0.5mm}b^0_k+\sum_{j=1}^h b_{jk} G_{ij} \tilde \bx_i \ba_j\hspace{-0.5mm}=\hspace{-0.5mm}(G_i*(\tilde \bx_i \bA))\bb_k+b^0_k, k\hspace{-0.5mm}=\hspace{-0.5mm}1,...,c
\] 
where $G_i$ is the $i$-th row of $\bG$, ``$*$'' is the elementwise multiplication, and $\tilde \bx_i=(1,\bx_i^T)$ is the $i$-th row of $\bX$.
Then, all responses can be written as
\[
\hat \bY=(\bG*(\bX\bA))\bB+\bb^0
\]

Adding the firing pattern $\bG$ to the loss function parameters gives 
\begin{equation}
L(\bA,\bB, \bb_0, \bG)=\frac{1}{N}\sum_{k=1}^c\sum_{i=1}^N \|\sum_{j=1}^h b_{jk} G_{ij} \tilde\bx_i \ba_j+b^0_k-y_{ik}\|^2
+\lambda (\|\bb^0\|^2+\|\bB\|_F^2+ \|\bA\|_F^2)).\label{eq:lossA}
\end{equation}
Let 
\begin{equation}
\bU_{jl}=\frac{1}{N}\sum_{i=1}^NG_{ij}G_{il}\tilde\bx_i\tilde\bx_i^T, \label{eq:U}
\end{equation}
which are 
$(d+1)\times (d+1)$ 
matrices that can be computed incrementally using batches. Let 
\begin{equation}
\bM=\begin{pmatrix}
\bM_{11} &\bM_{12} &... &\bM_{1h}\\
... &... &... &...\\
\bM_{h1} &\bM_{h2} &... &\bM_{hh} 
\end{pmatrix}\label{eq:M}
\end{equation}
be the 
$(d+1)h\times (d+1)h$ 
matrix with the cell $\bM_{jl}=\sum_{k=1}^c b_{jk}b_{lk}\bU_{jl}$. 

Also let 
\begin{equation}
\bc=(
\bv_{1}^T,...,\bv_{h}^T)^T, \bv_j=\frac{1}{N} \sum_{k=1}^c\sum_{i=1}^Nb_{jk} (y_i-b^0_k)G_{ij}\bx_i.\label{eq:C}
\end{equation}

The following Theorem describes the linear system of equations that need to be solved in order to find the critical points of the loss function  $L(\bA,\bB, \bb^0,\bG)$ w.r.t. $\bA$ analytically.

\begin{Theorem}
If the matrix $\bG$ from Eq. \eqref{eq:lossA} is fixed, the critical points with respect to $\bA$ of the loss function $L(\bA,\bB, \bb^0,\bG)$ from Eq. \eqref{eq:lossA} are solutions of the equation:
\begin{equation}
(\bM+\lambda \bI_{(d+1)h}) \ba=\bc, \label{eq:A}
\end{equation}
where $\ba=(\ba_1^T,...,\ba_h^T)^T$ is the matrix $\bA$ unraveled.
\end{Theorem}
\noindent{\em Proof.} The loss \eqref{eq:lossA} can also be written as
\begin{equation}
\begin{split}
L(\bA,\bB, b_0,\bG)=&\frac{1}{N}\sum_{k=1}^c\sum_{j,l=1}^h\ba_j^T b_{jk}b_{lk}(\sum_{i=1}^NG_{ij}G_{il}\bx_i\bx_i^T)\ba_l
+\frac{2}{N}\sum_{j=1}^h \sum_{k=1}^c \sum_{i=1}^Nb_{jk}(b^0_k-y_i)G_{ij}\bx_i^T\ba_j\\
&+\lambda (\bb^{0T}\bb^0+\sum_{k=1}^c\bb_j^T\bb_k+\ba^T\ba),
\end{split}
\end{equation}
therefore
\begin{equation}
L(\bA,\bB, b_0,\bG)=\hspace{-0.9mm}\sum_{j,l=1}^h\hspace{-0.5mm}\ba_j^T\hspace{-0.5mm}\left (\sum_{k=1}^c b_{jk}b_{lk}\bU_{jl}\right )\ba_l\hspace{-0.5mm}-\hspace{-0.5mm}2 \hspace{-0.5mm}\sum_{j=1}^h\hspace{-0.5mm}\bv_j^T\ba_j
+\lambda (\bb^{0T}\bb^0+\sum_{k=1}^c\bb_j^T\bb_k+\ba^T\ba).
\end{equation}

Then the loss function can be written using Eq. \eqref{eq:M} as
\begin{equation}
L(\ba,\hspace{-0.5mm}\bB, \hspace{-0.5mm}b_0,\hspace{-0.5mm}\bG)\hspace{-0.8mm}=\hspace{-0.8mm}\ba^T \bM \ba-2\bc^T\ba+\lambda (\bb^{0T}\bb^0\hspace{-0.5mm}+\hspace{-0.5mm}\sum_{k=1}^c\hspace{-0.5mm}\bb_j^T\bb_k+\ba^T\hspace{-0.8mm}\ba).
\end{equation}

This is a quadratic function in $\ba$ and by setting its gradient with respect to $\ba$ to zero we obtain Equation \eqref{eq:A}.$\qed$

\subsection{Computation Complexity}

The computation complexity of the analytic minimization algorithm ANMIN described in Algorithm \ref{alg:anmin} can be easily calculated as follows.

Computing the input $\bS$ for solving $\bB$ is $O(Ndh)$ and fitting $\bB$ is $O(h^3+ch^2)$. Computing each matrix $\bU_{ij}$ is $O(Nd^2)$, so computing $\bM$ is $O(N(dh)^2)$. 
Fitting $\bA$ is $O((dh)^3+c(dh)^2)$. Since there are a fixed number of iterations, the whole algorithm is $O((dh)^3+(N+c)(dh)^2)$, so it is linear in $N$ and $c$, but cubic in $d$ and $h$.

\begin{table}[ht]
\centering
\begin{tabular}{lccc}
Dataset &$N$ &$d$ &$h$ \\
\hline
SDF decoder &73,080 &2 &64\\
Abalone &4,177&7 &64\\
Bike-sharing &17,379 &13 &64 \\
Yearpred MSD&515,345&90 &64 \\
DAE&15,912&225 &32 \\
\hline
\end{tabular}
\vskip -3mm
\caption{The real datasets used in the experiments with their number of observations $n$ and feature dimension $d$. Also shown is the number of NN hidden nodes $h$ used in experiments.}
\label{tab:datasets}
 \vspace{-2mm}
\end{table}

\begin{figure}[t]
\centering
\includegraphics[width=0.5\linewidth]{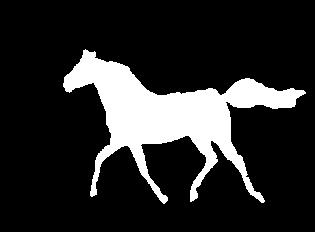}
\includegraphics[width=0.35\linewidth]{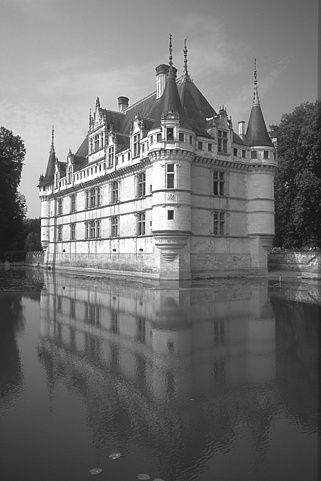}
\vspace{-1mm}
\caption{Left: the binary shape image whose signed distance transform was used to train a shape decoder. Right: the image used for training a denoising autoencoder (DAE). }\label{fig:image}
 \vspace{-1mm}
\end{figure}
\subsection{Datasets Used}

Real data experiments will be conducted on three datasets from the UC Irvine Machine Learning repository \cite{dua2019} and two datasets that were specially generated for two computer vision tasks. These five datasets are summarized in Table \ref{tab:datasets}.

From the UC Irvine Machine Learning repository \cite{dua2019} were used the abalone, bike-sharing and year prediction MSD datasets, all three being regression datasets.

The abalone dataset is about predicting the age of abalone from its physical measurements. It has 4177 observations and 7 features, so it is a low dimensional dataset.

The bike sharing dataset is about predicting the hourly and daily bike rental counts in a bike rental system. 
The hourly count was used, containing 17,379 observations and 14 features. 
The feature that counts the number of registered users was removed since it is very strongly correlated with the response, and the remaining 13 features were used for prediction.

The year Prediction MSD dataset is about predicting the release year of a song from audio features extracted from the song. It contains 515,345 observations and 90 features.

Another dataset was generated based on the single shape deep SDF task from \cite{park2019deepsdf}. 
It is about predicting the value of a signed distance transform (SDF) matrix from the pixel coordinates $(x,y)$.  
As input shape was used the first mask image of the Weizmann horse dataset \cite{borenstein2004combining}, shown in Figure \ref{fig:image}, left, from which the SDF was computed. 
Taking all pixels coordinates of the image as inputs and their corresponding SDF values as targets, a dataset with 73,080 observations and two features was obtained.

All these regression datasets are about predicting a single output. 
To evaluate the performance of ANMIN for predicting multiple outputs, the NN was used as a Denoising Autoencoder (DAE). 
In this case the desired outputs consist of natural image patches and the inputs are the same patches corrupted by noise.
The patches were extracted from an image from the Berkeley dataset \cite{MartinFTM01}, shown in Figure \ref{fig:image}, right. 
Using a step size of 3, from the image were extracted 15,912 overlapping patches of size $15\times 15$. Thus the data has 15,912 observations, 225 inputs and 225 outputs. The outputs are the original image patches, and the inputs are the same image patches corrupted with Gaussian noise with standard deviation $\sigma=10$.
\begin{figure}[h]
\vspace{-2mm}
\centering
\includegraphics[width=0.4\linewidth]{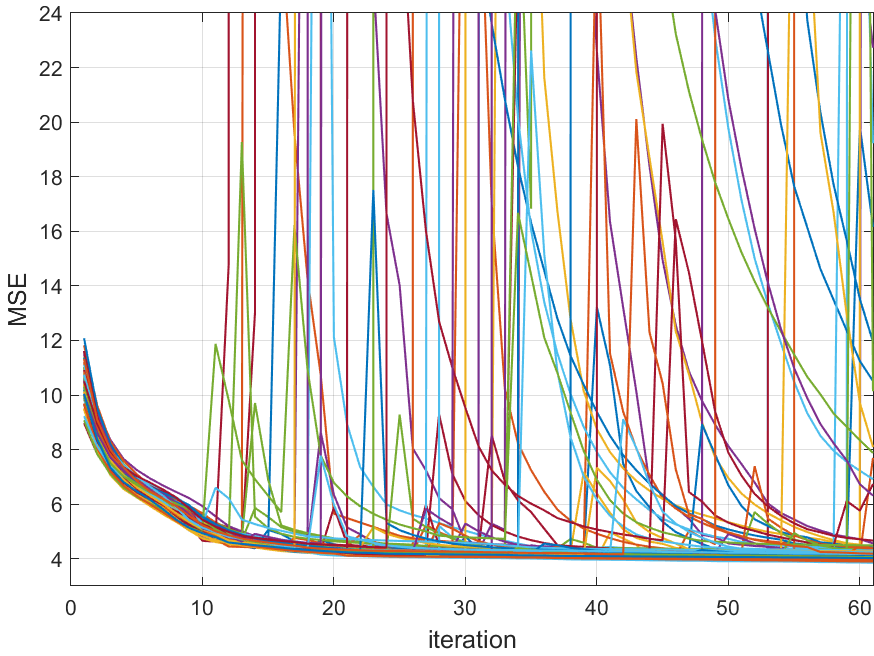}
\includegraphics[width=0.4\linewidth]{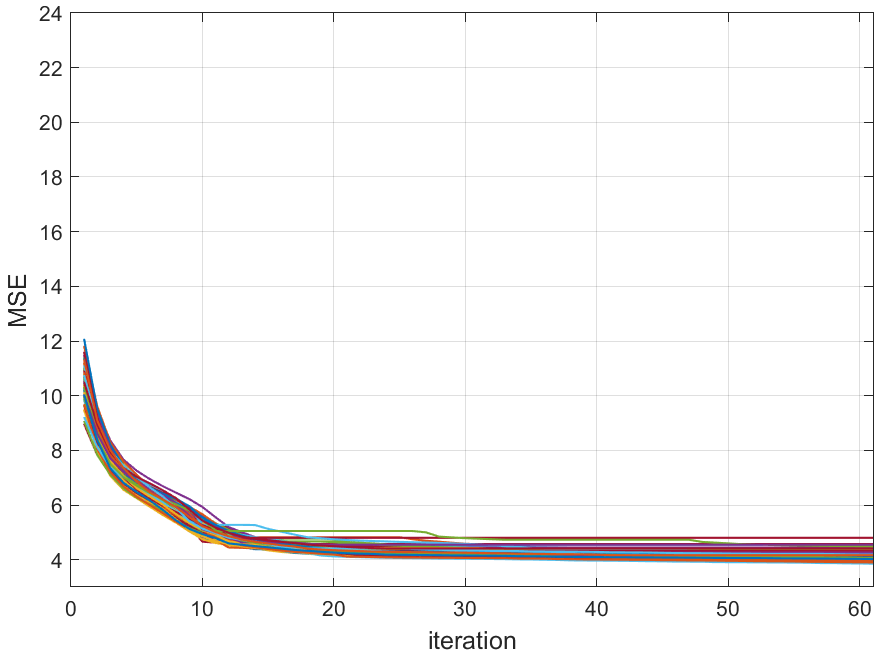}
\vspace{-2mm}
\caption{Training MSEs for LBFGS on 100 random splits of the abalone dataset. Left: the training MSEs have many places where LBFGS blows up. Right: considering the minimum train MSE obtained so far at each iteration alleviates the problem. }\label{fig:lossbf}
 \vspace{-4mm}
\end{figure}

\begin{table}[t]
\centering
\scalebox{0.97}{
\begin{tabular}{l|cccccc}
&\multicolumn{6}{c}{MSE train}  \\
Dataset &SGD &Adam &Adam-L &LBFGS &ANMIN  &ANMIN-L \\
\hline 
SDF decoder &22.43(5.29) &12.35(4.48) &9.46(2.49) &25.71(62.08) &{\bf 3.53}(1.05)&{\bf 3.45}(1.15) \\
Abalone &4.20(.08) &4.00(.09) &3.99(.09) &4.17(.16) &{\bf 3.94}(.10)&{\bf 3.95}(.09) \\
Bike-sharing &2422(100) &2465(181) &2477(193) &5710(4581) &{\bf 1457}(319) &{\bf 1555}(465) \\
YearPredMSD  &{\bf 71.67}(0.16) &{\bf 70.98}(.15)&{\bf 71.01}(.14) &74.51(2.55) &72.87(.18)&73.08(.21) \\
DAE &200.8(3.1)  &165.5(4.9) &163.4(4.5) &2990(9.2) &155.2(4.0)&{\bf 150.9}(2.7) \\
\hline
  &\multicolumn{6}{c}{MSE test}\\
Dataset  &SGD &Adam &Adam-L&LBFGS &ANMIN  &ANMIN-L \\
\hline 
SDF decoder &22.36(5.18)  &12.36(4.46)&9.48(2.49) &25.65(61.96) &{\bf 3.56}(1.06) &{\bf 3.48}(1.14)\\
Abalone &{\bf 4.45}(.41)  &5.05(1.89) &4.99(1.71)&{\bf 4.50}(.46) &{\bf 4.64}(.55) &{\bf 4.66}(.95)\\
Bike-sharing &2681(171) &2721(224)  &2737(239) &5850(4280) &{\bf 1714}(385) &{\bf 1808}(493)\\
YearPredMSD &{\bf 76.51}(0.53)  &{\bf 76.71}(.51)&{\bf 76.68}(.55)&77.17(1.29) &{\bf 76.88}(.54)&{\bf 76.82}(.49)\\
DAE   &207.1(6.1) &170.8(7.1) &168.7(6.1)&2988(36.8) &160.4(5.6) &{\bf 156.1}(4.8)\\

\hline 
&\multicolumn{6}{c}{$R^2$ train}  \\
Dataset &SGD  &Adam &Adam-L &LBFGS &ANMIN  &ANMIN-L \\
\hline 
SDF decoder &0.968(.008) &0.982(.007)&0.986(.004) &0.963(.089) &{\bf 0.995}(.002)&{\bf 0.995}(.002) \\
Abalone &0.596(.005) &0.615(.005)&0.615(.006)&0.598(.014) &{\bf 0.621}(.007)&{\bf 0.620}(.007) \\
Bike-sharing &0.926(.003) &0.925(.006)&0.925(.006)&0.826(.139)&{\bf 0.956}(.010) &{\bf 0.953}(.014) \\
YearPredMSD &0.400(.001) &{\bf 0.406}(.001)&{\bf 0.406}(.001)&0.376(.021) &0.390(.001)&0.388(.002) \\
DAE &0.933(.001)  &0.945(.002) &0.945(.002) &0.000(.000)&0.948(.001) &{\bf 0.950}(.002) \\
\hline
&\multicolumn{6}{c}{$R^2$ test}  \\
Dataset &SGD &Adam &Adam-L &LBFGS &ANMIN  &ANMIN-L \\
\hline 
SDF decoder &0.968(.008) &0.982(.006) &0.986(.004) &0.963(.090) &{\bf 0.995}(.002)&{\bf 0.995}(.002)\\
Abalone &{\bf 0.573}(.028) &0.515(.178)&0.521(.164)&{\bf 0.568}(.035) &{\bf 0.555}(.049)&{\bf 0.554}(.079)\\
Bike-sharing &0.919(.005) &0.918(.006)&0.917(.007)&0.822(.130)&{\bf 0.948}(.012)&{\bf 0.945}(.015)\\
YearPredMSD &{\bf 0.360}(.003)  &0.359(.003) &0.359(.003)&0.355(.011) &0.357(.003)&0.358(.003)\\
DAE &0.931(.002) &0.943(.002) &0.944(.002) &0.000(.000)&0.946(.002)&{\bf 0.948}(.002)\\
\hline

\end{tabular}
}
\vskip -2mm
\caption{Real data experiments. Average final train and test MSE values and $R^2$ (all with std) on five datasets. All results are obtained from 100 random splits into 80\% train and 20\% test. The best results and the ones not significantly worse ($p>0.01$) are shown in bold.}
\label{tab:results}
\vspace{-3mm}
\end{table}
For all datasets, one hundred random splits of the data into an 80\% train set and 20\% test set were generated.

A 2-layer NN with $h$ hidden nodes and ReLU or Leaky ReLU activation was trained on the training parts of the random splits using different training algorithms, and then tested on the corresponding test sets. 
The number of hidden nodes was $h=64$ except for the DAE data where it was $h=32$ to obtain a low dimensional representation.

\subsection{Methods Compared}

 The proposed ANMIN algorithm was compared with the stochastic gradient descent algorithm, the Adam optimizer \cite{kingma2014adam} and the LBFGS algorithm \cite{liu1989limited}. 
The main focus are SGD and Adam, because the LBFGS blows up to values as large as $10^{26}$ many times during optimization, as one could see in Figure \ref{fig:lossbf}, left for the abalone dataset, where 100 training losses were plotted. 
For this reason, for the LBFGS algorithm, for each run the model that obtained the smallest training MSE was extracted and used for testing. 
The same approach was used for ANMIN (lines 17-20 of Algorithm \ref{alg:anmin}).

ANMIN and Adam were experimented with two activation functions: ReLU ($\alpha=0$) and leaky ReLU ($\alpha=0.1$), while LBFGS and SGD were experimented only with ReLU. 
The methods with leaky ReLU are called ANMIN-L and Adam-L in experiments.
\begin{figure*}[ht]
\centering
\begin{tabular}{ccc}
\includegraphics[width=0.28\linewidth]{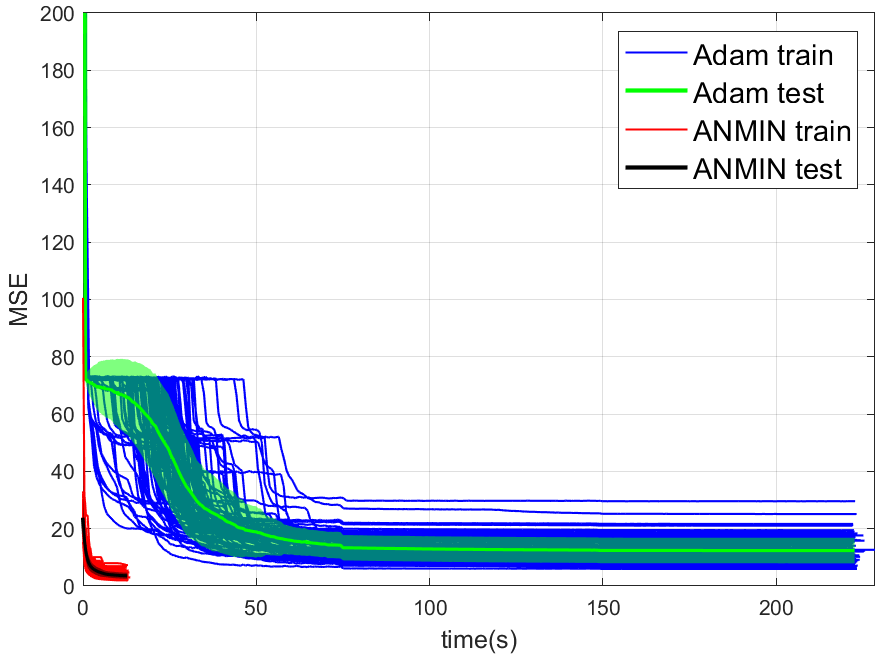}
&\includegraphics[width=0.28\linewidth]{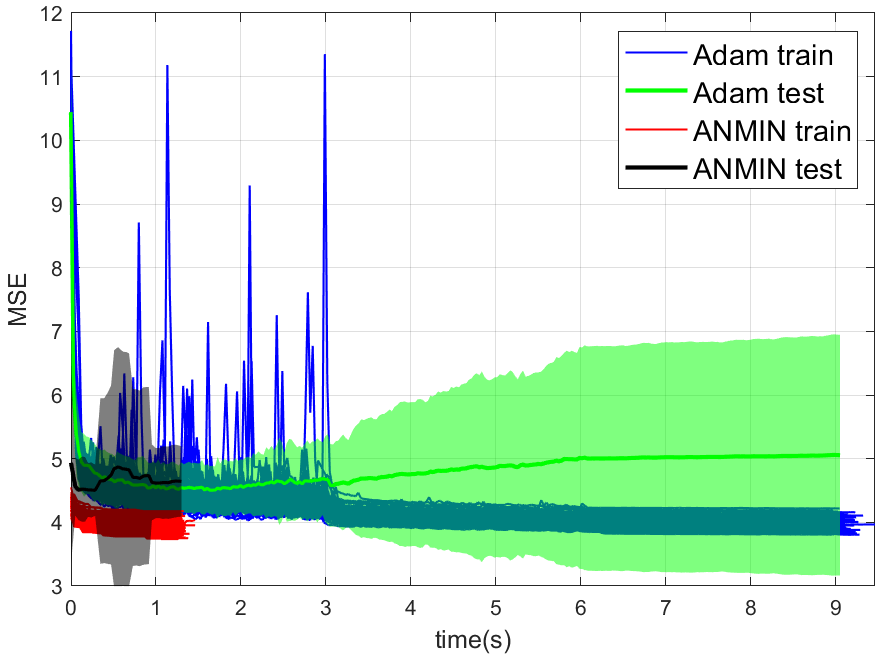}
&\includegraphics[width=0.28\linewidth]{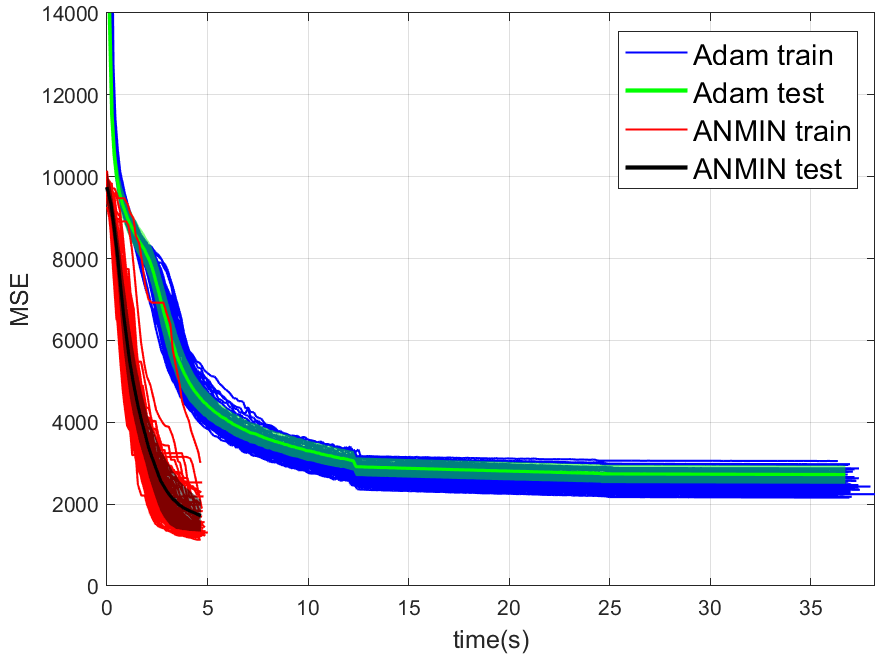}\vspace{-2mm}\\
\includegraphics[width=0.28\linewidth]{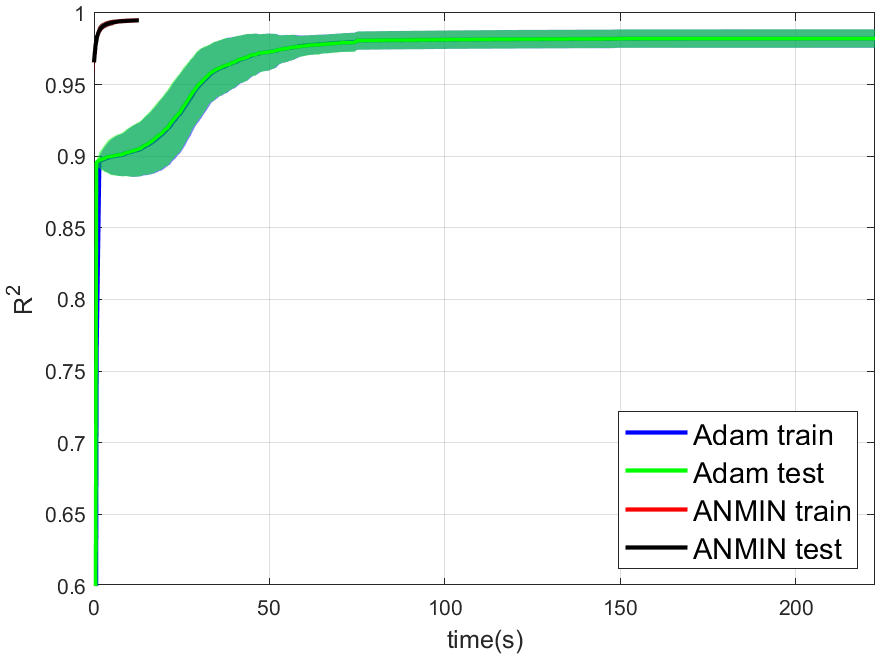}
&\includegraphics[width=0.28\linewidth]{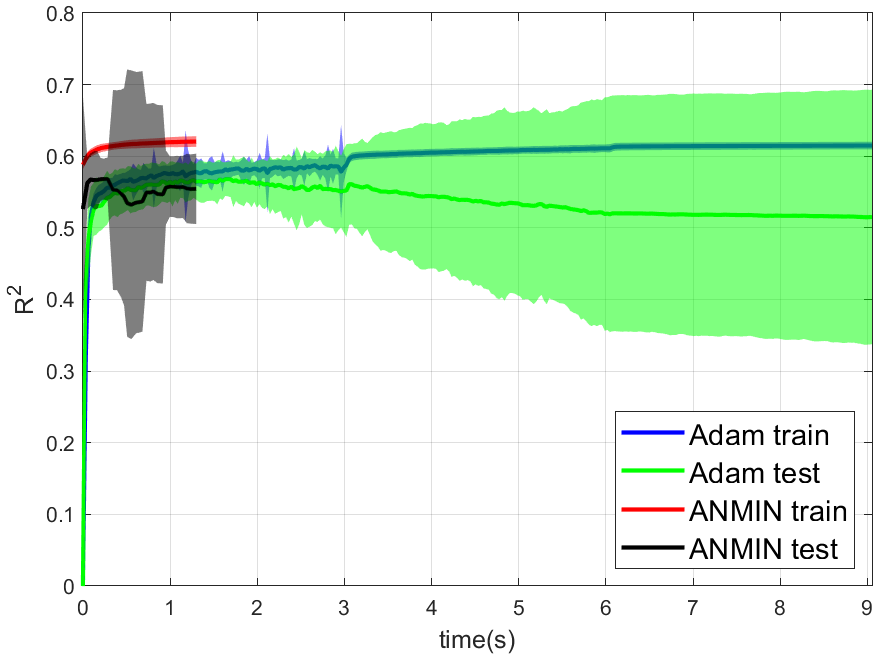}
&\includegraphics[width=0.28\linewidth]{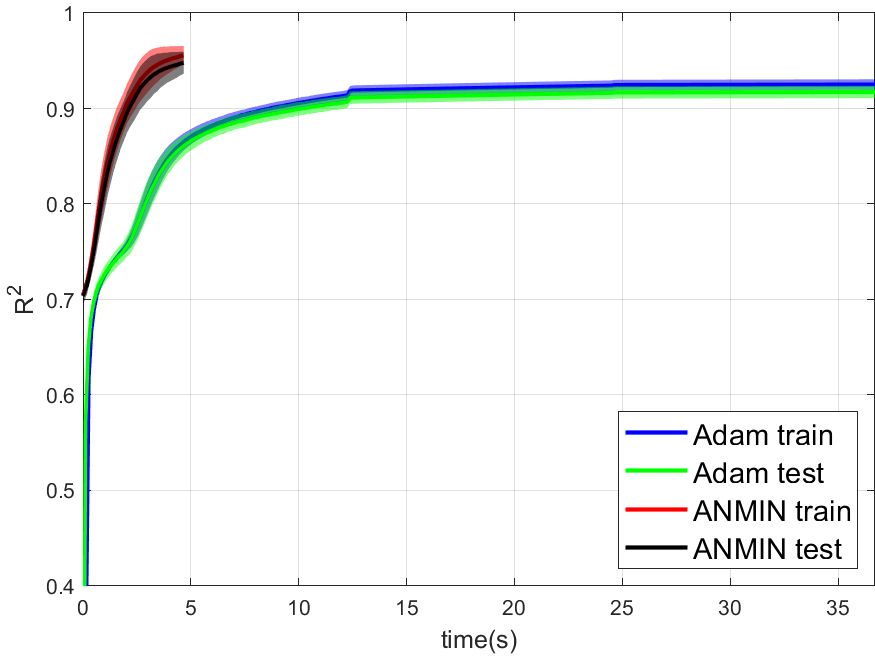}\vspace{-2mm}\\
SDF decoder  &abalone &bike-sharing 
\end{tabular}
\begin{tabular}{cc}
\includegraphics[width=0.28\linewidth]{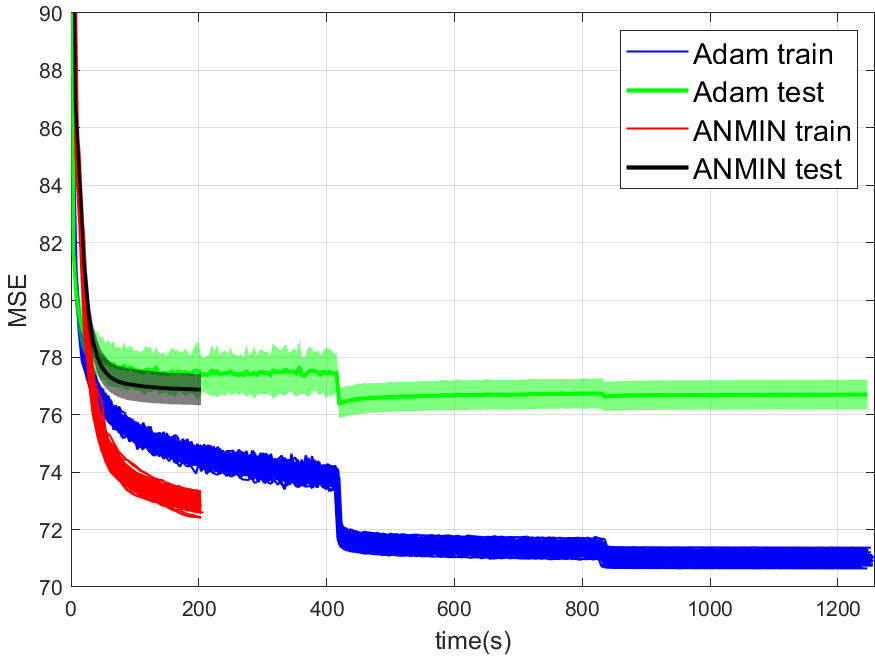}
&\includegraphics[width=0.28\linewidth]{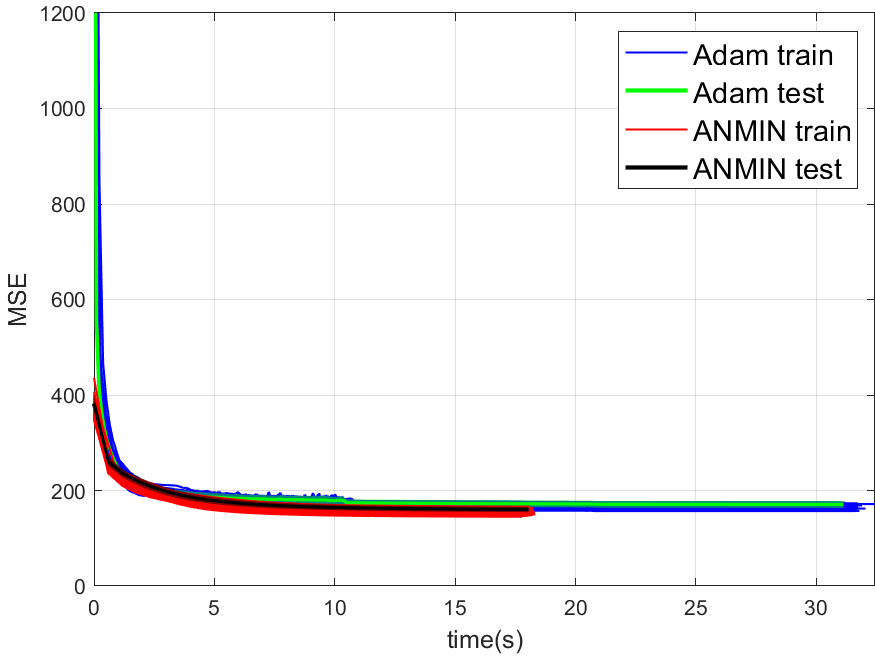}\vspace{-1mm}\\
\includegraphics[width=0.28\linewidth]{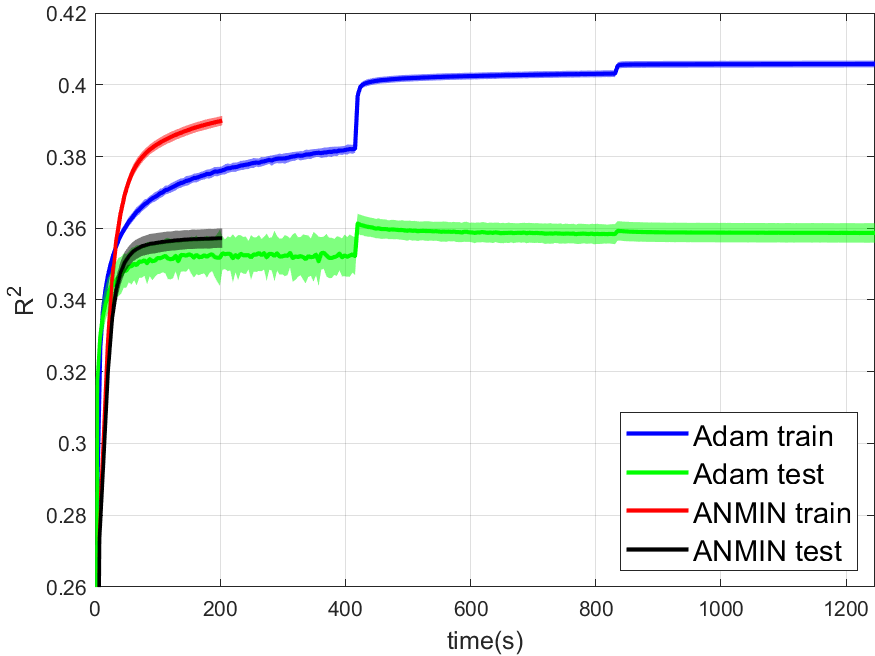}
&\includegraphics[width=0.28\linewidth]{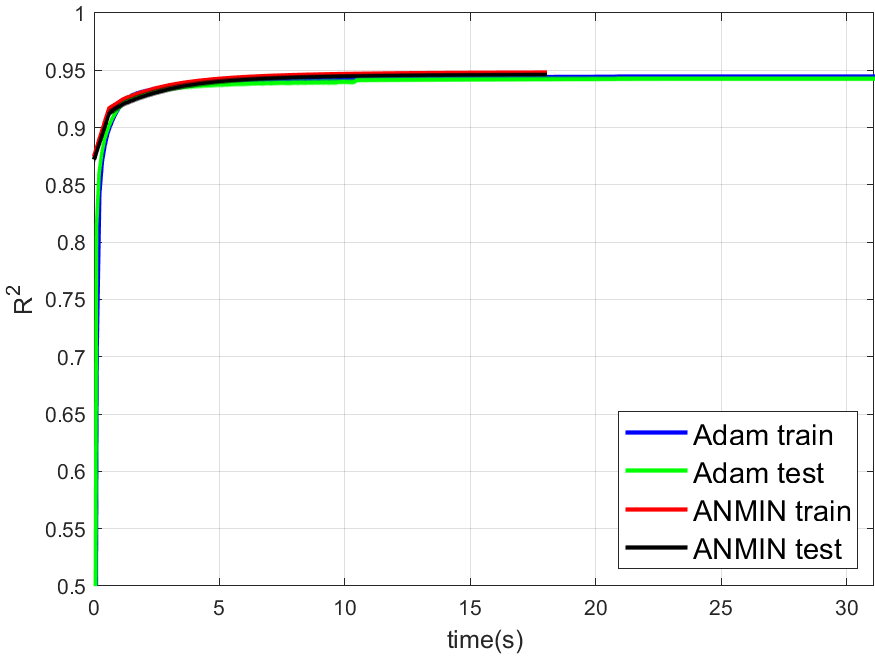}\vspace{-1mm}\\
Year pred MSD &DAE\\
\end{tabular}
\vspace{-2mm}
\caption{MSE and average $R^2$ vs time (seconds) of 100 runs of training the NN with ReLU activation using the Adam, and ANMIN optimizers. Also plotted are the mean test MSEs and $R^2$ with standard deviation. 
} \label{fig:mseAdam}
\vspace{-5mm}
\end{figure*}

\subsection{Implementation Details}

All algorithms were implemented in PyTorch. 
The experiments were conducted on a Core I7 computer with 32Gb RAM and GeForce 3080 GPU.
The code 
is available at \url{https://github.com/barbua/ANMIN/}. 

The SGD and Adam were trained for 300 epochs. 
The initial learning rate for Adam was 0.03, while for SGD it was tuned for each dataset individually to avoid the loss blowing up and to obtain a small final loss value. 
The learning rate was decreased by a factor of 10 after every 100 epochs. 
The ANMIN algorithm was run for 30 iterations. 
The LBFGS was run for 60 iterations with a learning rate of 0.01.

For Adam and SGD, a batch size of 256 was used except for Year Prediction MSD, where a batch of 2048 was used. 
For LBFGS a batch size of 100,000 was used.
For ANMIN, a batch size of 256 was used. 
The batch size is used in the ANMIN to accumulate the matrices $\bU_{jl}$ from Eq. \eqref{eq:U}, so different batch sizes result in identical solutions, and we have observed that the computation time is approximately the same for batch sizes between 128 and 2048.
The value of the shrinkage parameter $\lambda$ for ANMIN was $\lambda=0.001$.

\begin{figure*}[ht]
\centering
\begin{tabular}{ccc}
\includegraphics[width=0.28\linewidth]{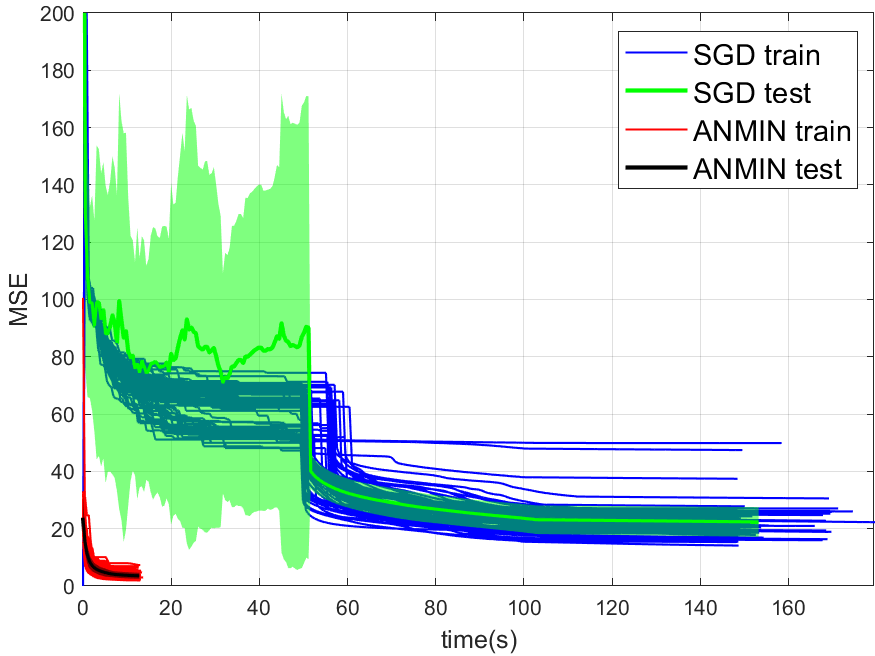}
&\includegraphics[width=0.28\linewidth]{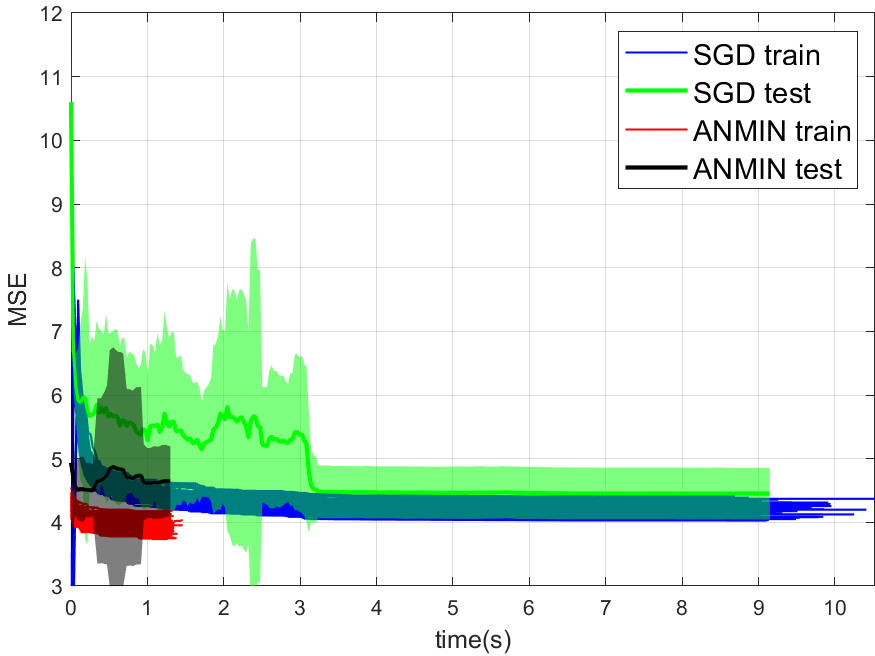}
&\includegraphics[width=0.28\linewidth]{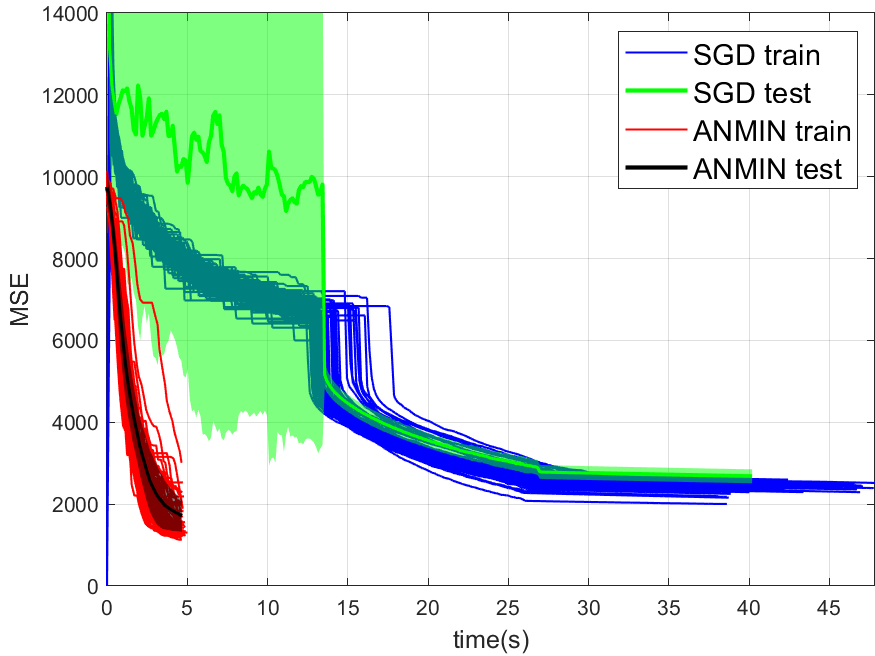}\\
\includegraphics[width=0.28\linewidth]{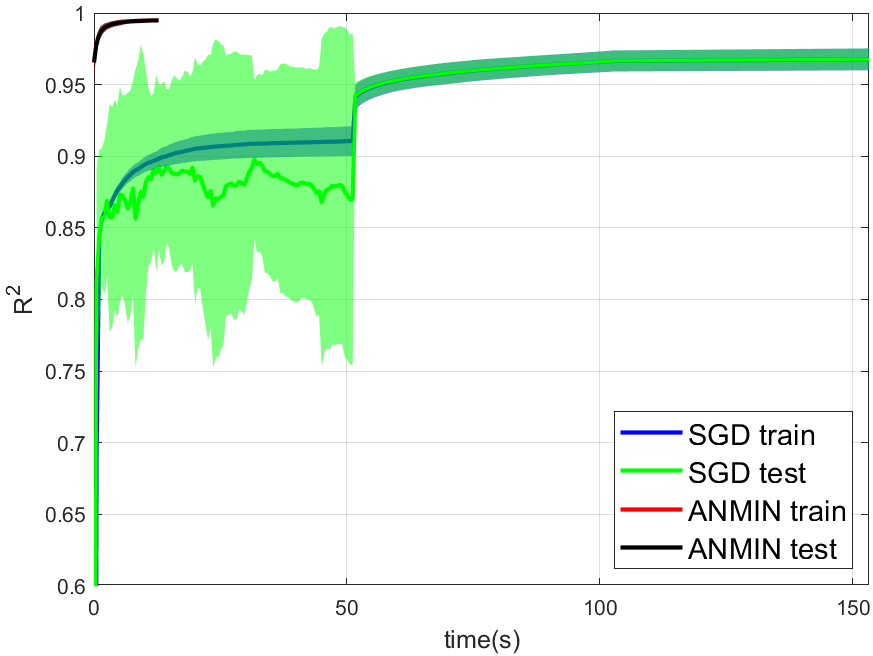}
&\includegraphics[width=0.28\linewidth]{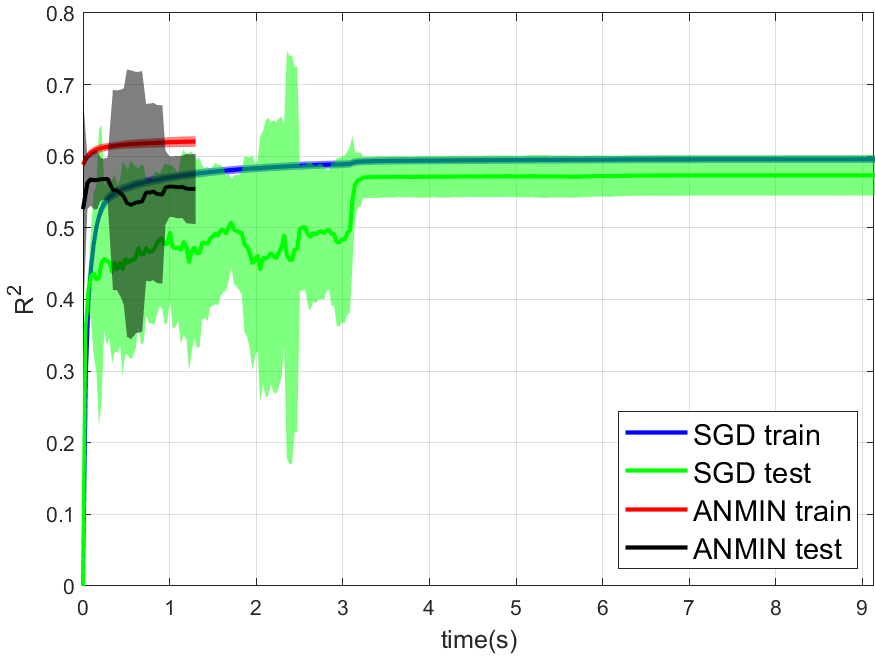}
&\includegraphics[width=0.28\linewidth]{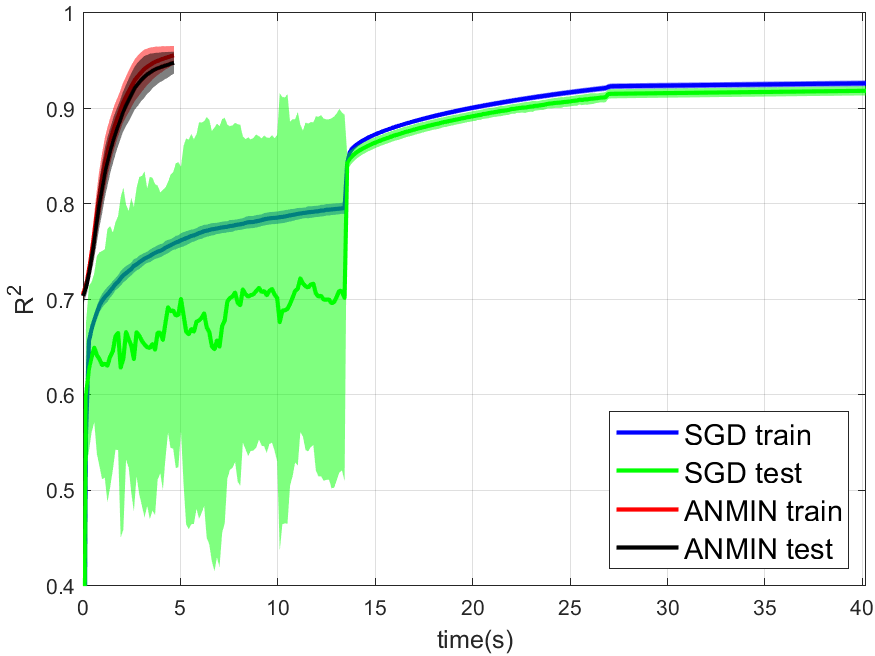}\vspace{-2mm}\\
SDF decoder  &abalone &bike-sharing 
\end{tabular}
\begin{tabular}{cc}
\includegraphics[width=0.28\linewidth]{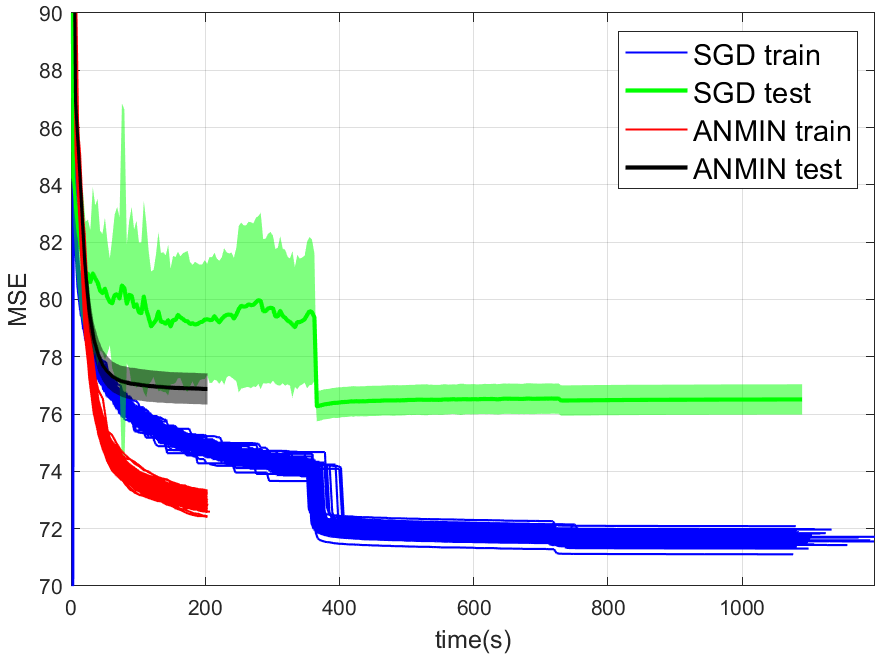}
&\includegraphics[width=0.28\linewidth]{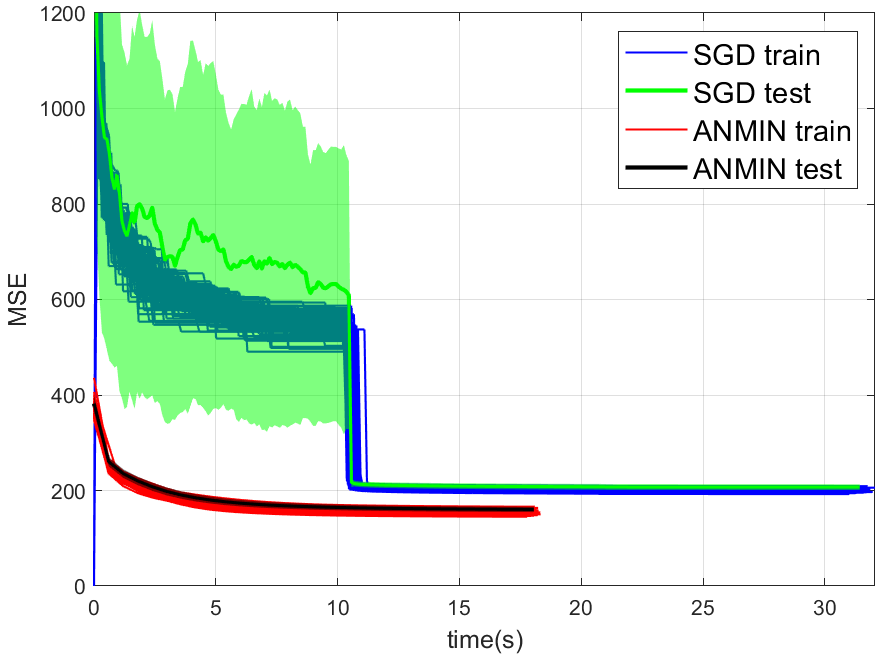}\\
\includegraphics[width=0.28\linewidth]{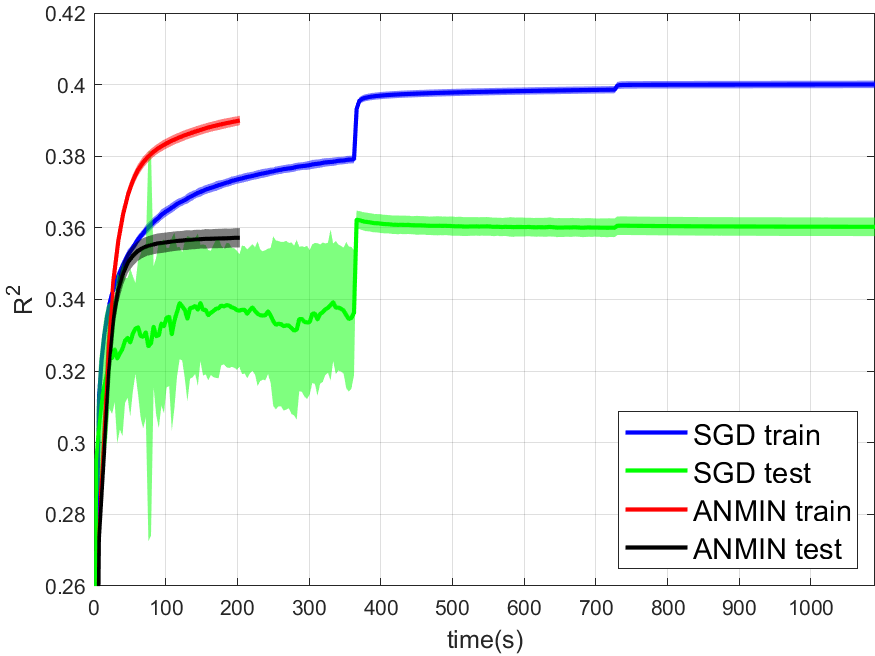}
&\includegraphics[width=0.28\linewidth]{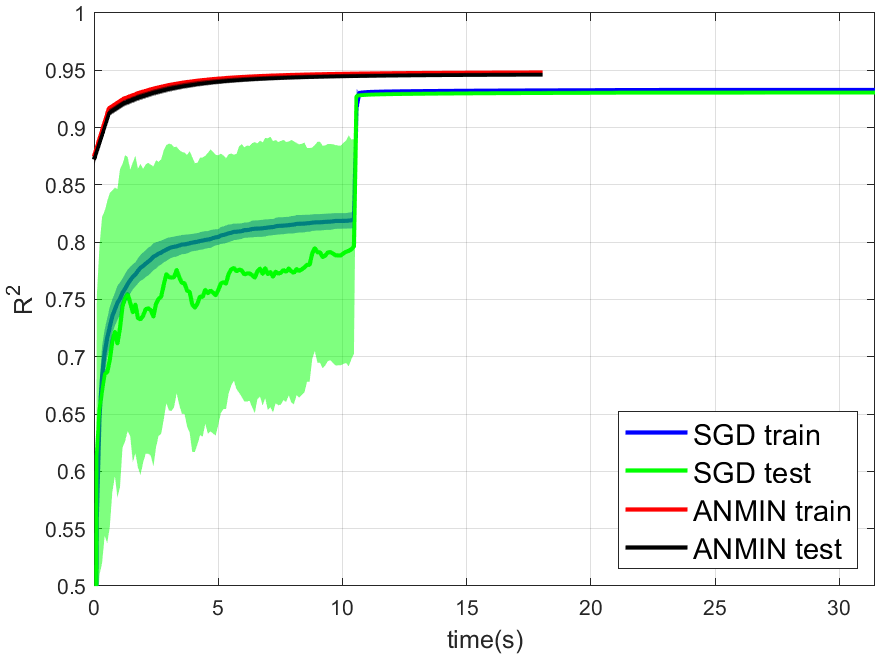}\vspace{-1mm}\\
Year pred MSD &DAE\\
\end{tabular}
\vspace{-2mm}
\caption{MSE and average $R^2$ vs time (seconds) of 100 runs of training the NN with ReLU activation using the SGD and ANMIN optimizers. Also plotted are the mean test MSEs and $R^2$ with standard deviation. 
} \label{fig:mseSGD}
\vspace{-5mm}
\end{figure*}

\section{Results and Discussion} \label{sec:results}

Experiments are conducted on multiple real datasets to obtain insight about when the ANMIN method works well compared to regular gradient-based training and when not. 
We will see that the advantages of ANMIN are more clear when the input dimension is small relative to the number of hidden neurons.
Then, simulations will be conducted on a nonlinear dataset where the input dimension and the number of observations can be modified as desired.

\subsection{Real Data Experiments}
\begin{figure*}[ht]
\centering
\begin{tabular}{ccc}
\includegraphics[width=0.28\linewidth]{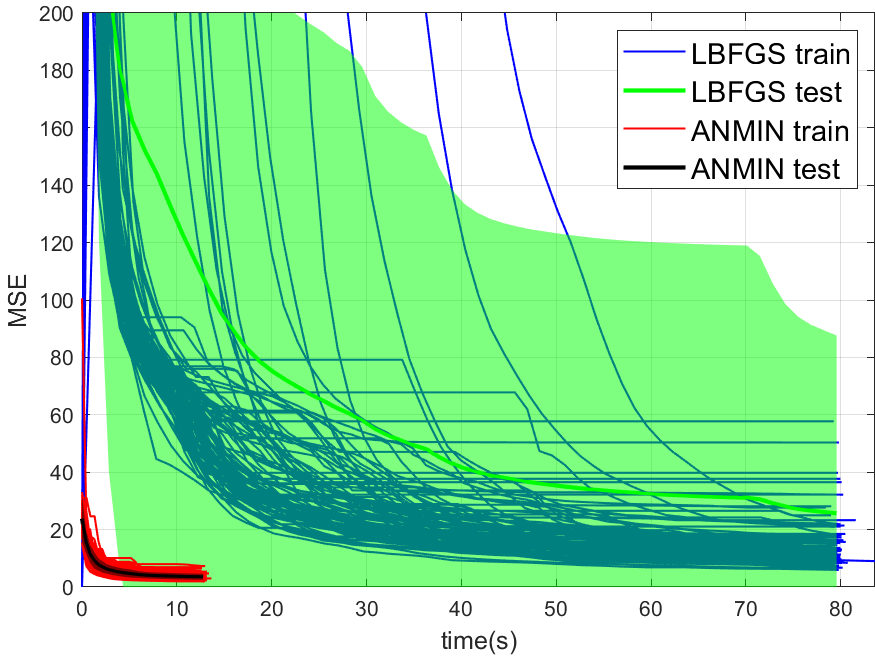}
&\includegraphics[width=0.28\linewidth]{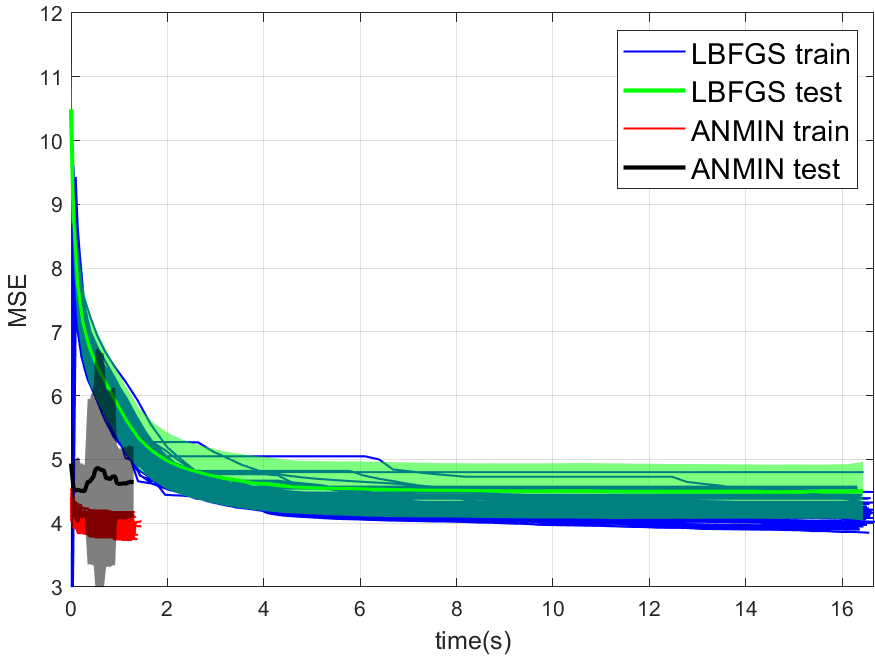}
&\includegraphics[width=0.28\linewidth]{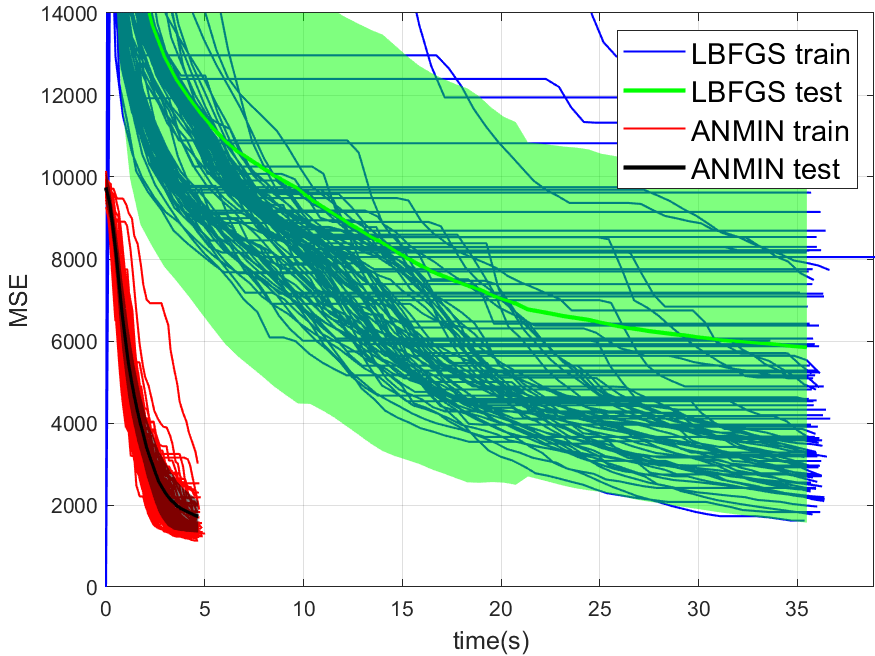}\\
\includegraphics[width=0.28\linewidth]{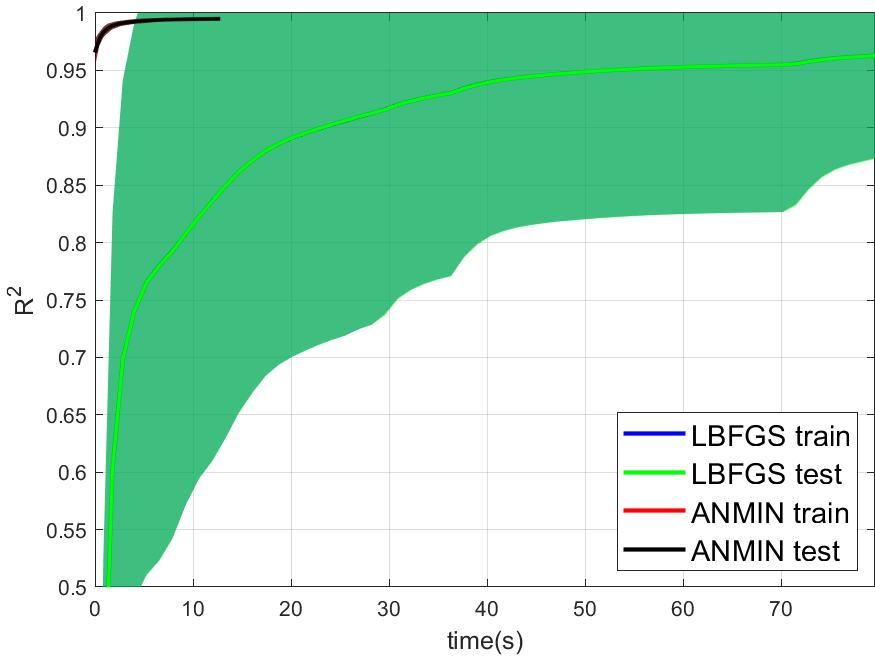}
&\includegraphics[width=0.28\linewidth]{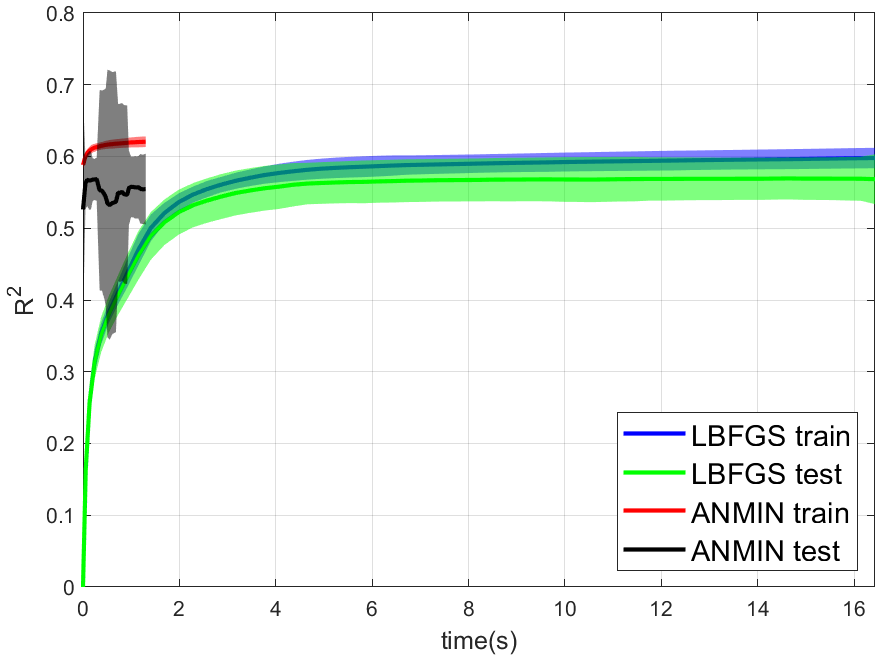}
&\includegraphics[width=0.28\linewidth]{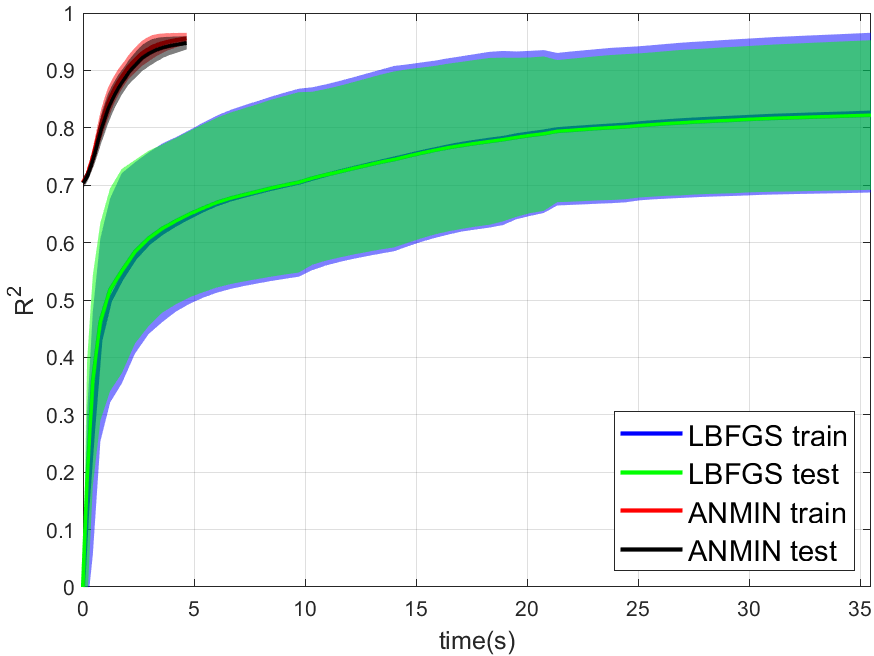}\vspace{-2mm}\\
SDF decoder  &abalone &bike-sharing 
\end{tabular}
\begin{tabular}{cc}
\includegraphics[width=0.28\linewidth]{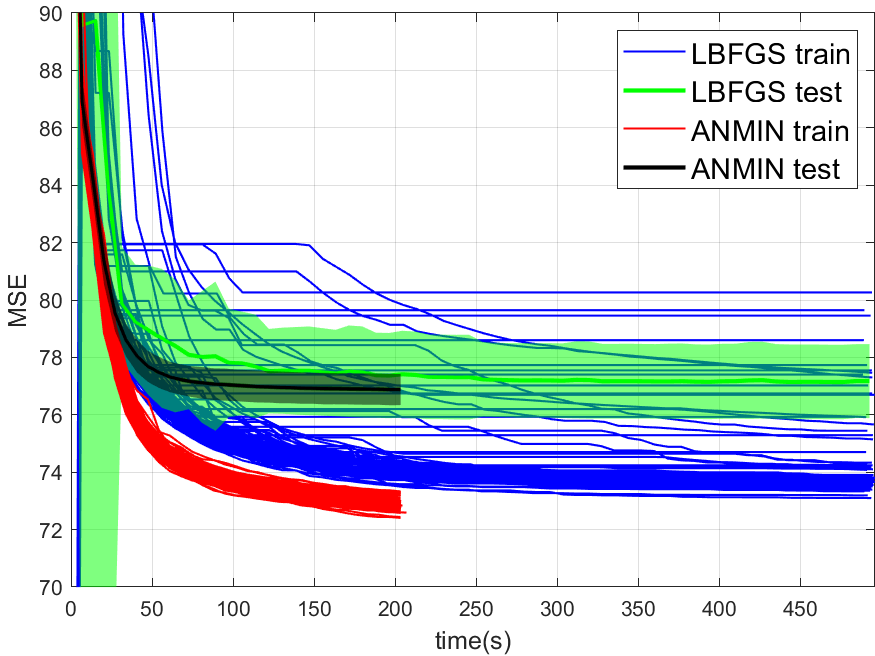}
&\includegraphics[width=0.28\linewidth]{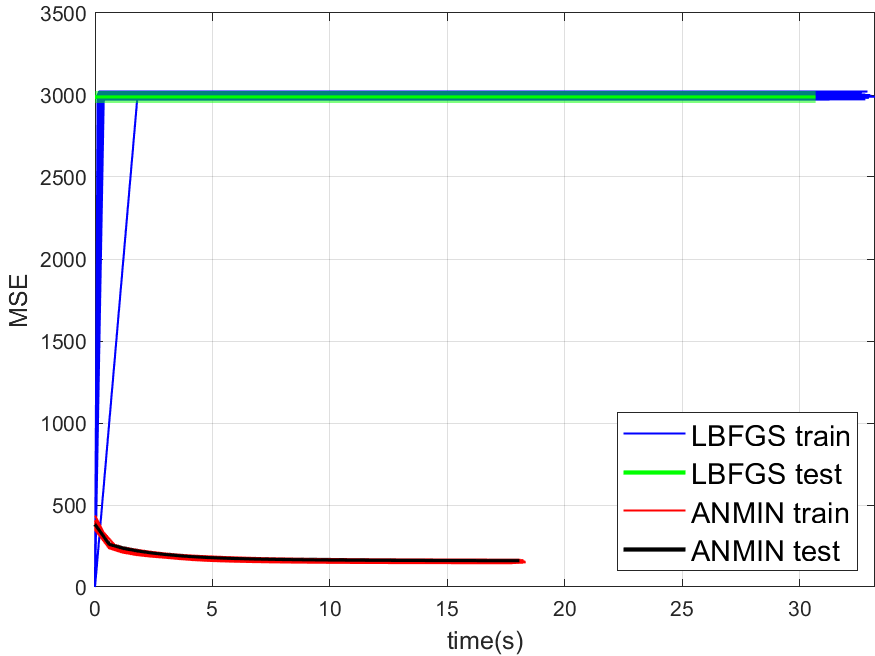}\\
\includegraphics[width=0.28\linewidth]{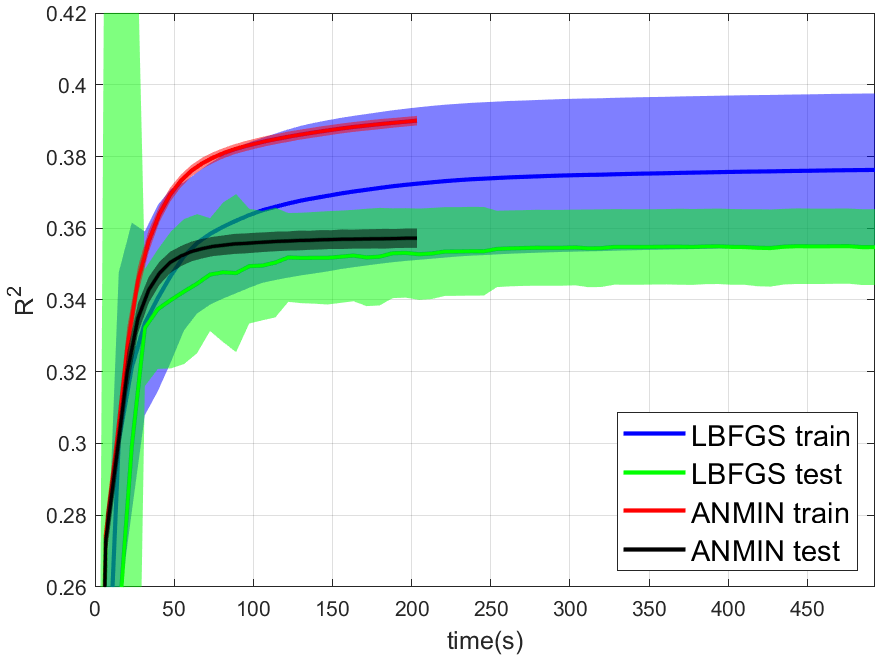}
&\includegraphics[width=0.28\linewidth]{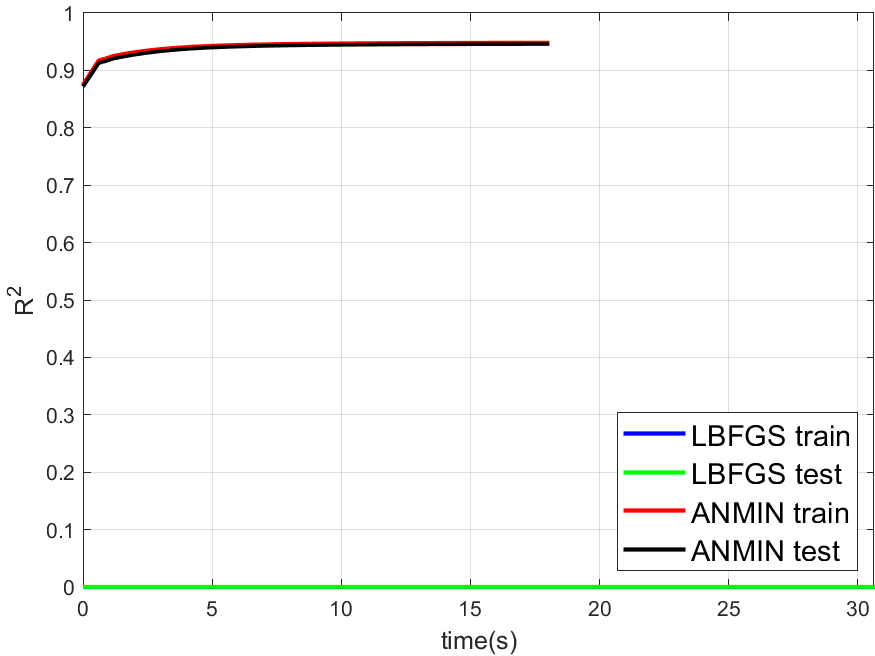}\vspace{-1mm}\\
Year pred MSD &DAE\\
\end{tabular}
\vspace{-2mm}
\caption{MSE and average $R^2$ vs time (seconds) of 100 runs of training the NN with ReLU activation using the LBFGS and ANMIN optimizers. Also plotted are the mean test MSEs and $R^2$ with standard deviation. 
} \label{fig:mseBF}
\vspace{-5mm}
\end{figure*}

The results as training and test MSEs and training and test $R^2$ are collected in Table \ref{tab:results}. 
The best result and the results that obtained a $p$-value $p>0.01$ based on a paired $t$-test significance comparison with the best result are shown in bold.

In Figures \ref{fig:mseAdam},  \ref{fig:mseSGD},  \ref{fig:mseBF} are plotted the training MSE loss and average $R^2$ curves for the ANMIN method vs Adam, SGD and LBFGS methods methods respectively (with ReLU activation), on all 100 random splits. 
Also plotted are the mean test MSE and $R^2$ curves with their standard deviation.
The test MSEs and $R^2$ for ANMIN and LBFGS are based on the model with the smallest training MSE loss.

Looking at the MSE loss values in Figures \ref{fig:mseAdam},  \ref{fig:mseSGD},  \ref{fig:mseBF}, and Table \ref{tab:results}, one could see that the ANMIN method obtains significantly lower train loss values than Adam, SGD and LBFGS on four out of the five datasets. 
It obtains significantly lower test MSEs than Adam and LBFGS on four datasets, and than SGD on three datasets. 
Moreover, it is never significantly outperformed on the test set by either SGD, Adam or LBFGS on any of the five datasets. 
The LBFGS algorithm performs very well only on abalone and is decent on Year Prediction MSD. 
It performs very poorly on the other three datasets.

Looking at the $R^2$ values from Figures \ref{fig:mseAdam},  \ref{fig:mseSGD},  \ref{fig:mseBF}, and Table \ref{tab:results}, we that the ANMIN method obtains significantly higher train $R^2$ values than SGD, Adam and LBFGS on four out of the five datasets. 
It obtains significantly higher test $R^2$ than Adam or LBFGS on four datasets, and than SGD on three datasets.
ANMIN is significantly outperformed by Adam and SGD on one dataset.

\begin{figure*}[ht]
\centering
\begin{tabular}{ccc}
\includegraphics[width=0.3\linewidth]{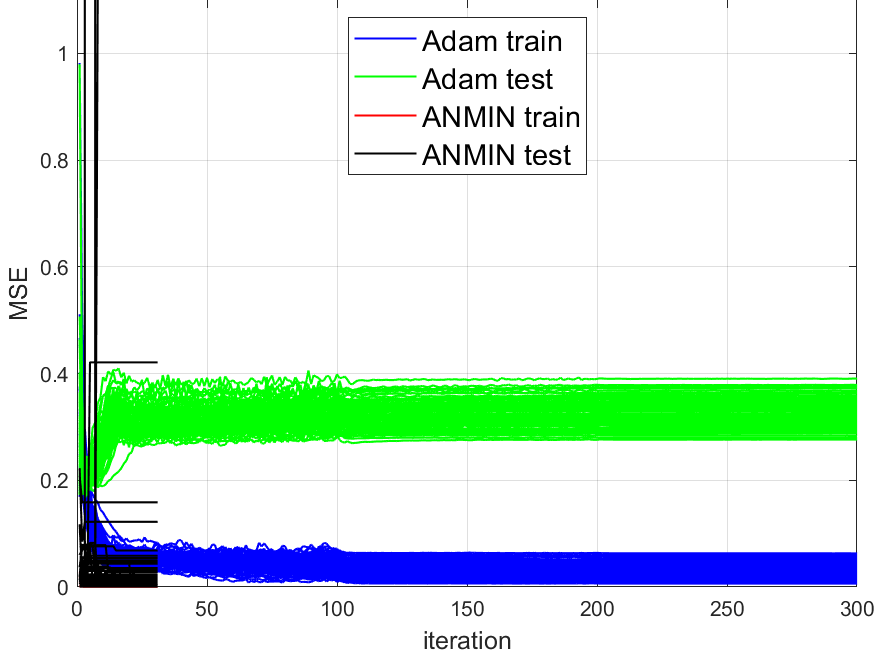}
&\includegraphics[width=0.3\linewidth]{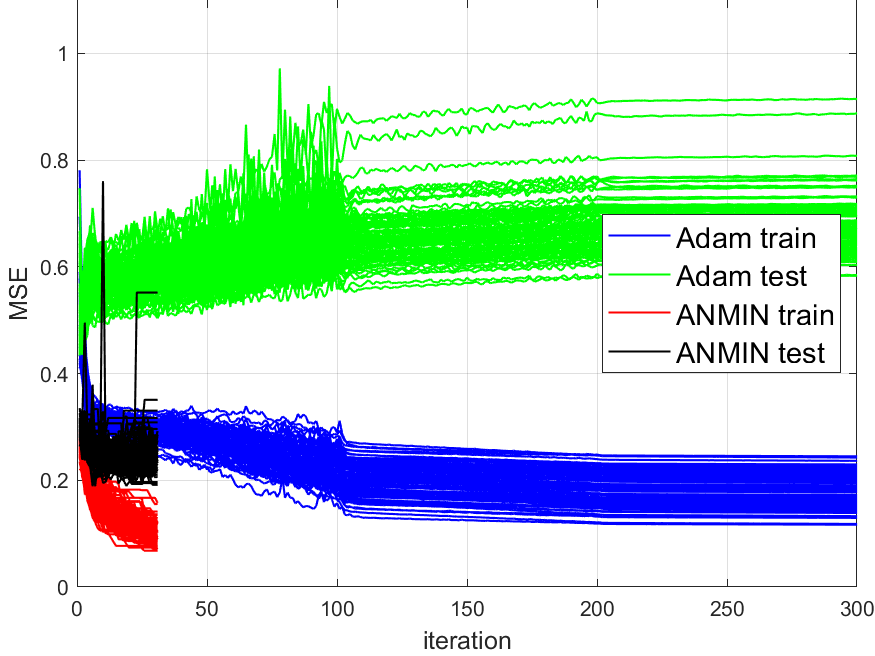}
&\includegraphics[width=0.3\linewidth]{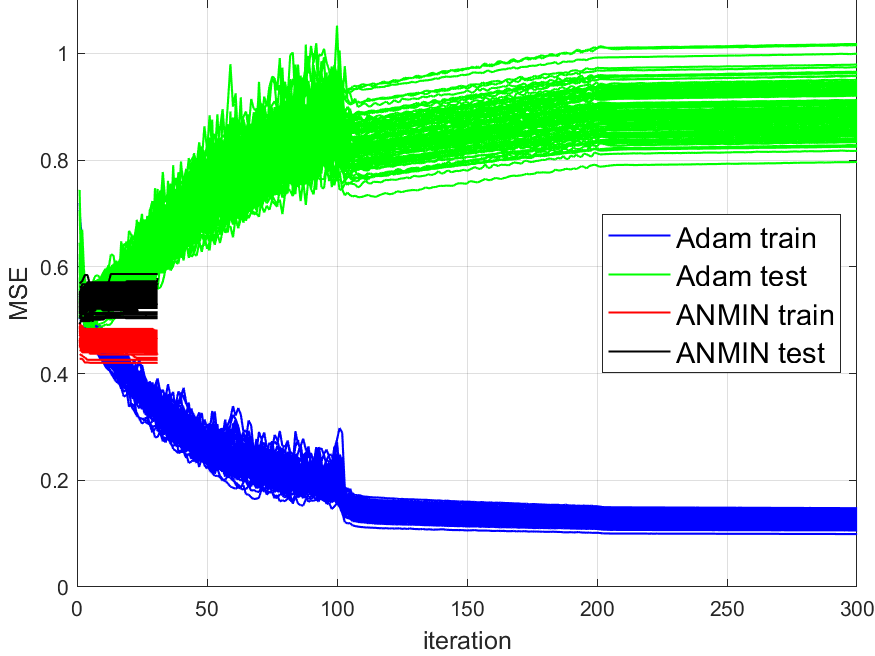}\vspace{-1mm}\\
$N=1,000,d=1$ &$N=1,000,d=3$  &$N=1,000,d=10$ 
\end{tabular}
\begin{tabular}{cc}
\includegraphics[width=0.3\linewidth]{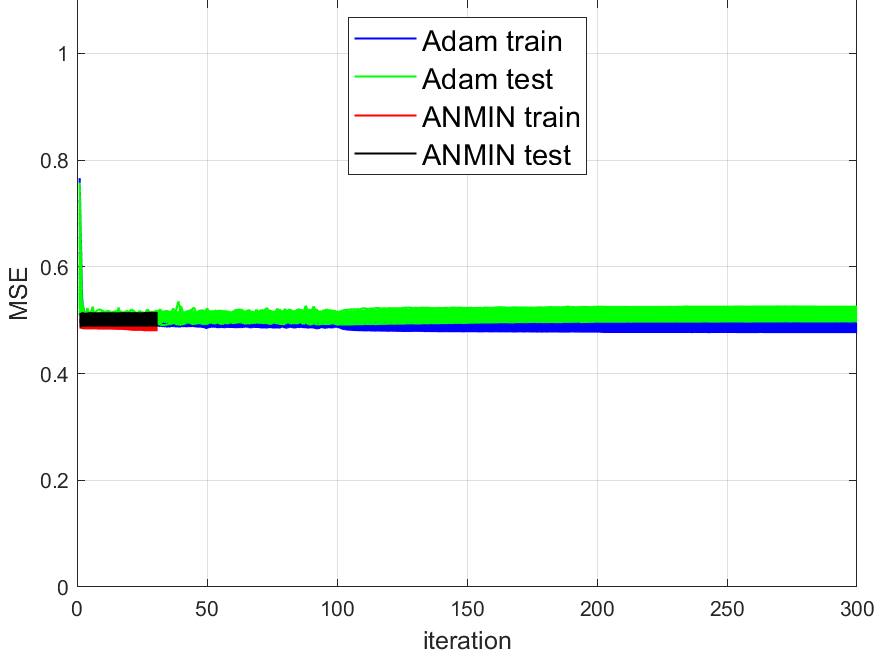}
&\includegraphics[width=0.3\linewidth]{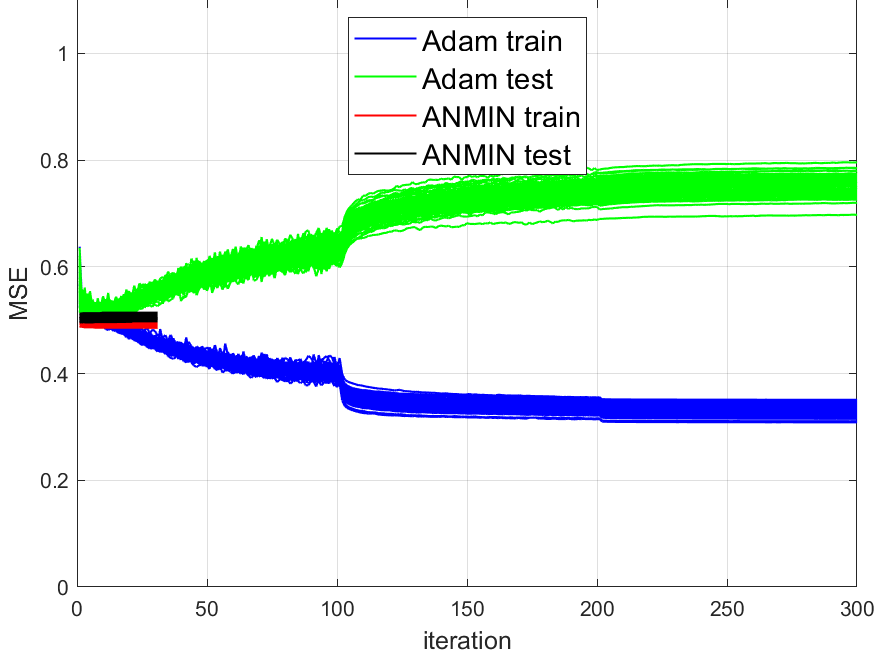}\vspace{-1mm}\\
$N=10,000,d=10$ &$N=10,000,d=30$
\end{tabular}
\vspace{-2mm}
\caption{Simulation experiments. MSEs of 100 runs of training a 64 hidden node NN using 300 epochs of the Adam optimizer or 30 iterations of ANMIN (Algorithm \ref{alg:anmin}).} \label{fig:plot_sin}
\vspace{-5mm}
\end{figure*}

What do the datasets where the ANMIN method outperforms Adam and SGD have in common?
One could notice that the one-output datasets in question all have low dimensional inputs, $d\leq 13$. 
Thus, the real data experiments seem to support the hypothesis that ANMIN has an advantage over standard gradient based optimization when the input dimension of the data (hence the number of parameters in the hidden layer) is small. 
This hypothesis will be further explored in the simulations Section \ref{sec:simulations}.
In that case, it seems that SGD and Adam have a hard time exploring the low dimensional parameter space  and finding a deeper local optimum.
Indeed, on the smallest dimensional dataset, which is the SDF decoder, ANMIN's MSE was about three times smaller than SGD's and Adam's. 
On abalone and bike sharing, the other two datasets with small numbers of features, ANMIN again significantly outperformed Adam and SGD on the training set, but SGD somehow generalized better on the test set for abalone.

The LBFGS algorithm did not do well on all datasets except for abalone, and in almost all cases its training MSE values were so high that it was the worst of all methods evaluated.
 
\begin{table}[h]
\centering
\begin{tabular}{lcccccc}
Dataset &SGD  &Adam &Adam-L &LBFGS &ANMIN &ANMIN-L\\
\hline
SDF decoder &153 &223&224 &79.6 &{\bf 12.8}&19.3\\
Abalone &9.1 &9.1&9.1&16.4 &1.3&{\bf 0.9}\\
Bike-sharing &40.2  &36.7&36.0 &35.5 &4.7&{\bf 3.2}\\
Yearpred MSD &1089 &1246&1084&493 &{\bf 203}&208\\
DAE &31.4 &31.1&31.4&30.7 &{\bf 18.1}&{\bf 18.1}\\
\hline
\end{tabular}
\caption{Computation times (seconds) for the experiments from Table \ref{tab:results}.}\label{tab:times}
\end{table}

The computation times for the methods evaluated are shown in Table \ref{tab:times}. 
Comparing computation times, one could see from Figure \ref{fig:mseAdam} that a train loss value as large as the final Adam loss is obtained by ANMIN in 5-10 times less time on the three low dimensional datasets, and more iterations bring even more improvements to the loss values.
Even the total computation times (30 ANMIN iterations vs. 300 Adam) show a computation advantage for the ANMIN method, in the range of 2 times faster for the DAE to more than 7 times for the three low dimensional datasets.
\begin{figure*}[ht]
\centering
\begin{tabular}{ccc}
\includegraphics[width=0.3\linewidth]{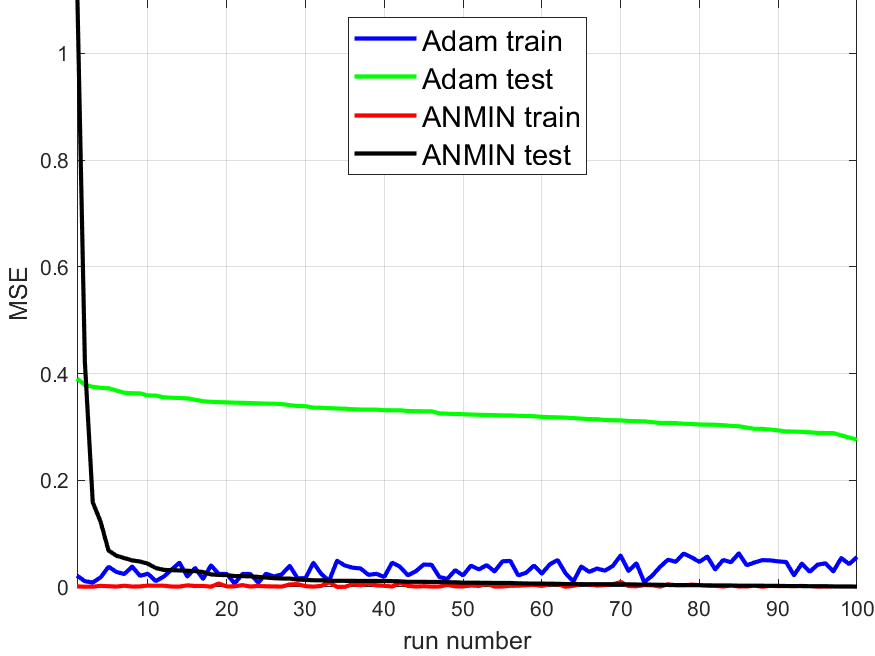}
&\includegraphics[width=0.3\linewidth]{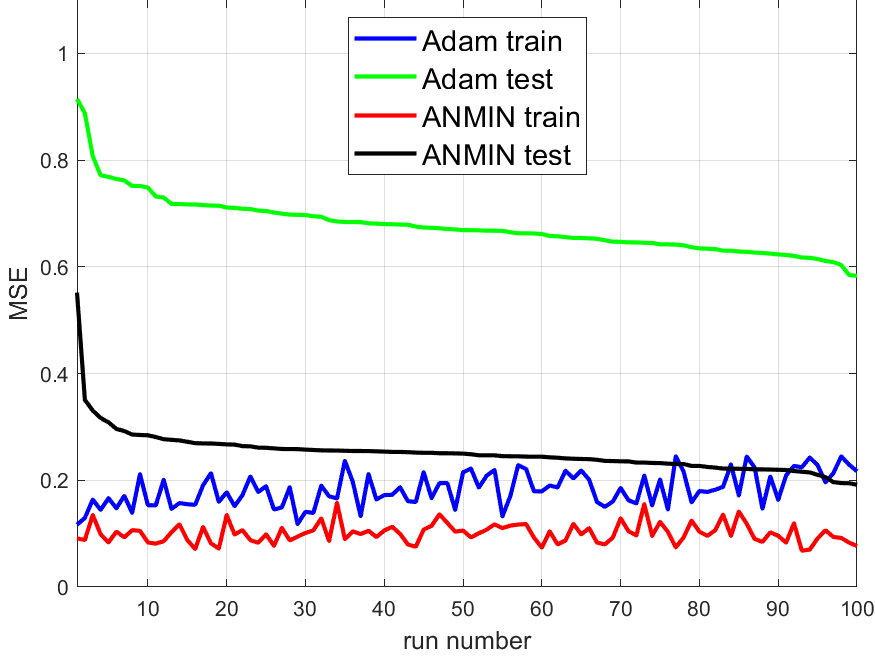}
&\includegraphics[width=0.3\linewidth]{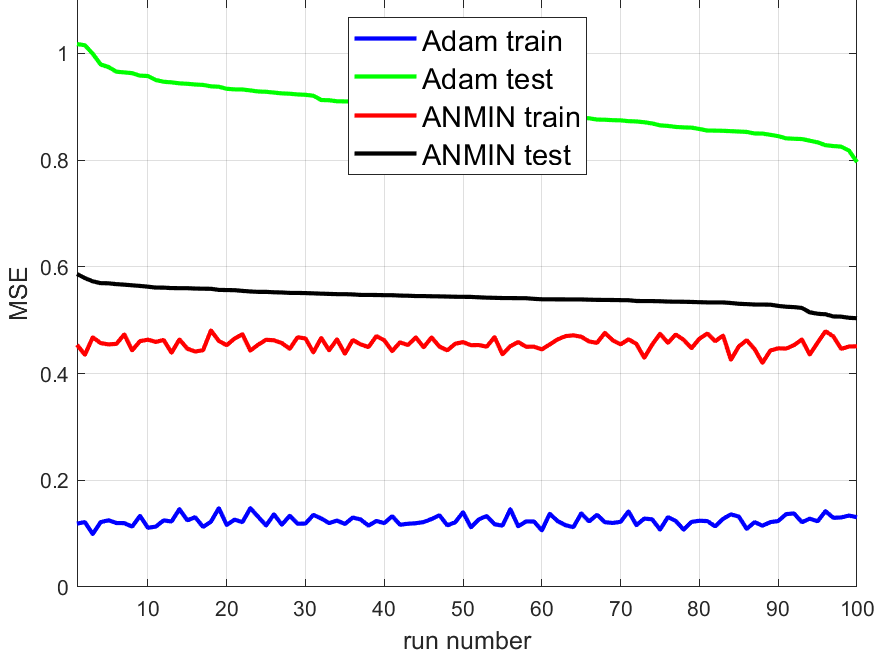}\vspace{-1mm}\\
$N=1,000,d=1$ &$N=1,000,d=3$  &$N=1,000,d=10$ 
\end{tabular}
\begin{tabular}{cc}
\includegraphics[width=0.3\linewidth]{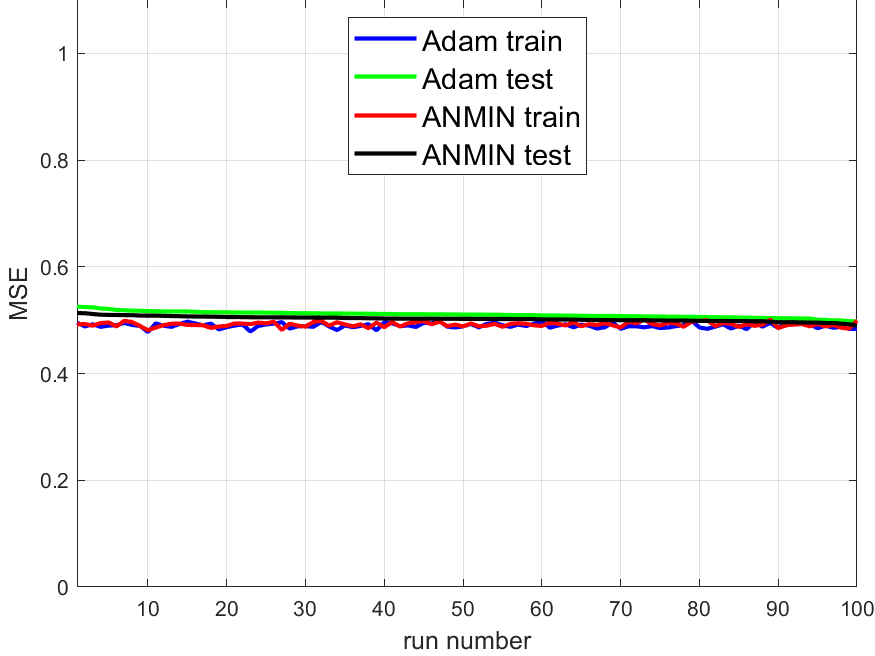}
&\includegraphics[width=0.3\linewidth]{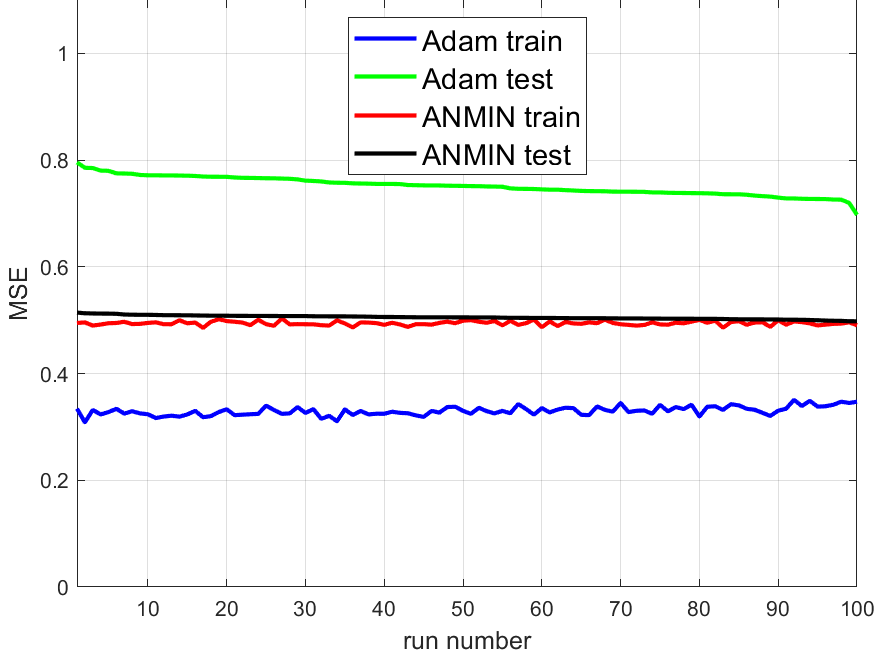}\vspace{-1mm}\\
$N=10,000,d=10$ &$N=10,000,d=30$
\end{tabular}
\vspace{-2mm}
\caption{Simulation experiments. Final train and test MSEs obtained by the Adam and ANMIN over the 100 runs, sorted in decreasing order of the test MSE.} \label{fig:plot_sin_srt}
\vspace{-5mm}
\end{figure*}

These experiments indicate that the ANMIN method might have an advantage in some cases, e.g. when the input dimension $d$ is small. This hypothesis is further investigated in the next section.

\subsection{Simulations} \label{sec:simulations}

In this section, simulations will be conducted to test the hypothesis that the advantage of ANMIN vs Adam is more clear when the input dimension is small.

The simulated data has $\bx\in \RR^d$ sampled from ${\cal N}(0,\bI_d)$ and $y=\sin(\|\bx\|_2^2)$. The dimension $d$ was taken from $d\in \{1,3,10,30\}$.
A number of $N\in \{1000,10000\}$ observations were sampled for training and an equal number for testing.

The model is a 2-layer NN with $h=64$ hidden nodes and ReLU activation. 
It is worth noting that the model does not fit the data well when $d$ is large.

The model was trained with 30 iterations of ANMIN (Algorithm \ref{alg:anmin}).
For comparison, it was also trained with with 300 iterations of the Adam optimizer \cite{kingma2014adam} starting with learning rate $\eta=0.03$, which was reduced by 10 every 100 iterations. The batch size was 256.

The plots of the MSE vs iteration for 100 consecutive runs for different combinations of $n$ and $d$ are shown in Figure \ref{fig:plot_sin}. 
For a better visualization, in Figure \ref{fig:plot_sin_srt} are plotted the train and test MSEs obtained by the two methods over the 100 runs, where the runs are sorted by decreasing test MSE for each method independently. 
This type of display could allow one to observe any trend between the train an test MSEs for each method. 
It would also allow one to get a better idea of the distribution of the test MSEs.

One could see that for small $d\in \{1,3\}$, the ANMIN can obtain both train and test loss values smaller than Adam. 
For $d\in \{10,30\}$, Adam obtains a smaller training loss, but it overfits more than ANMIN. 
In all situations, the ANMIN method obtains smaller test MSE values.

\begin{figure*}[ht]
\centering
\begin{tabular}{cccc}
\includegraphics[width=0.24\linewidth]{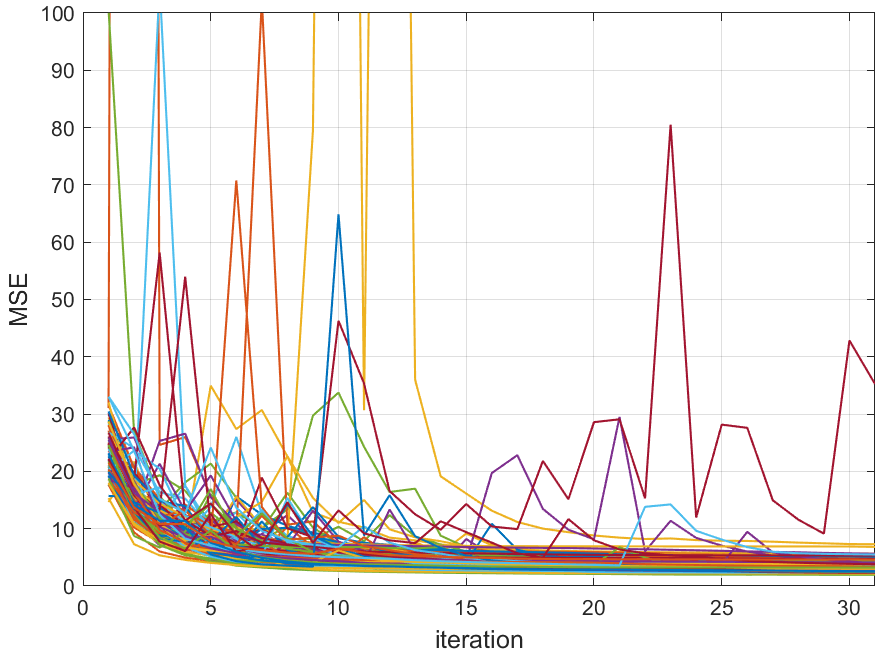}
&\includegraphics[width=0.24\linewidth]{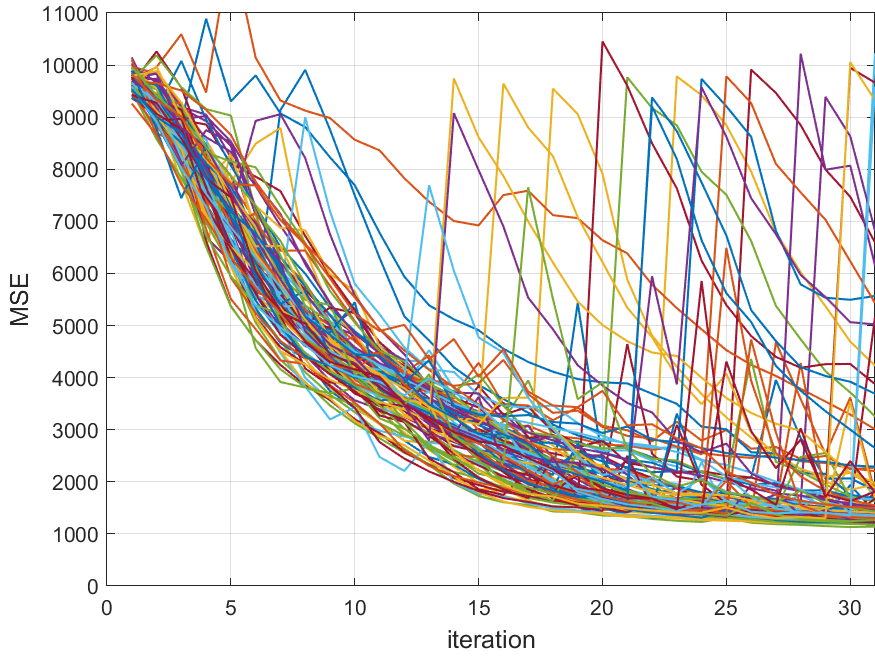}
&\includegraphics[width=0.24\linewidth]{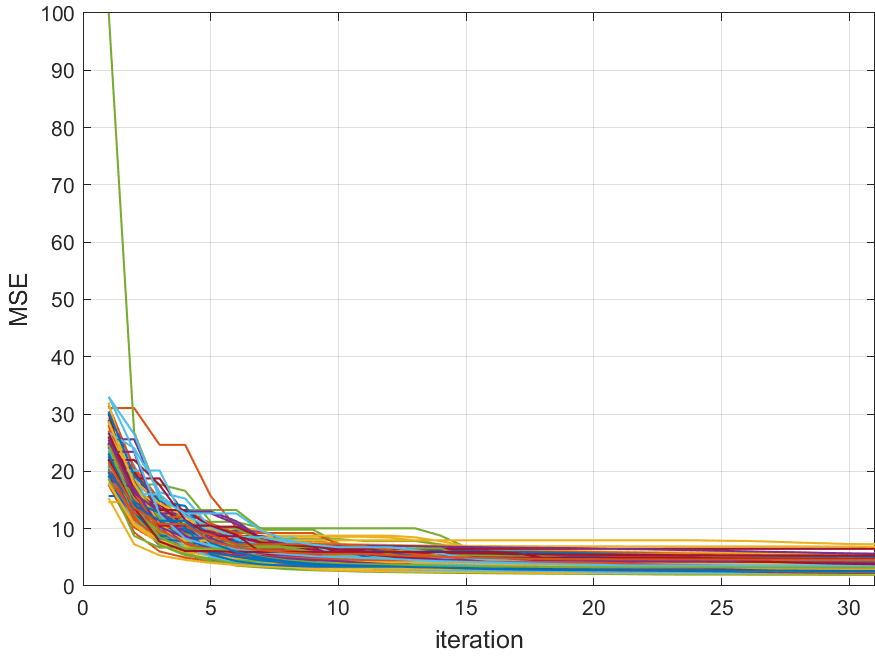}
&\includegraphics[width=0.24\linewidth]{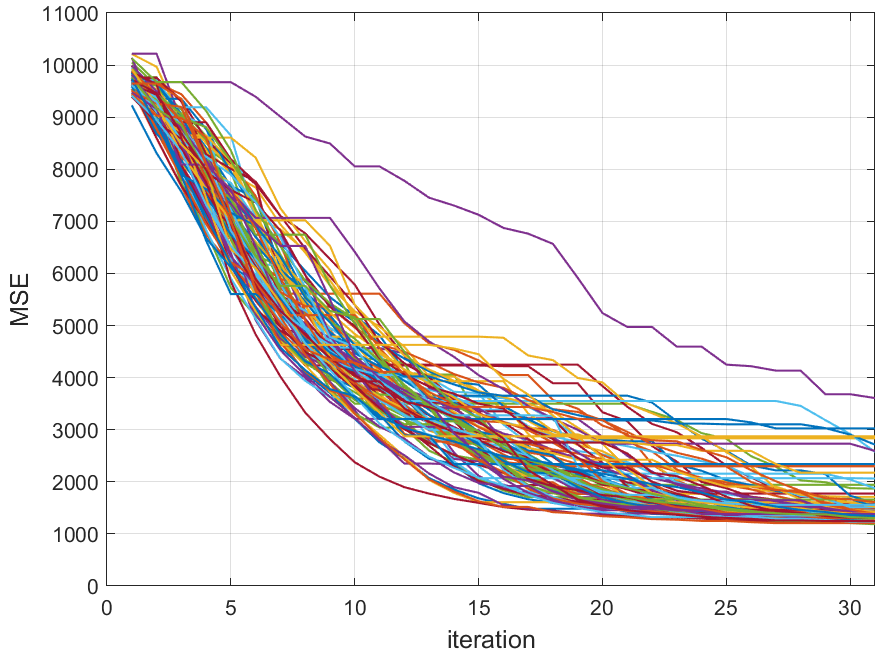}\\
SDF decoder &bike sharing  &SDF decoder &bike sharing\\
\end{tabular}
\vspace{-2mm}
\caption{MSE plots vs time (seconds) of 100 runs of training a 64 hidden node NN using 30 iterations of ANMIN (Algorithm \ref{alg:anmin}). Left two plots: the actual MSE values encountered at each iteration. Right two plots: the minimum MSE obtained so far at each iteration.} \label{fig:loss_bike}
 \vspace{-2mm}
\end{figure*}

\subsection{Ablation Study}

This section investigates how the design decisions made for Algorithm \ref{alg:anmin} affect its performance.
The design decisions that will be investigated are: 
\begin{itemize}
\item Whether to use the minimum loss value encountered during the iterations instead of the final loss of the algorithm.
\item When the linear system \eqref{eq:A} is degenerate, whether to stop, reinitialize $\bA$ randomly or continue with a pseudoinverse.
\end{itemize}

The ablation experiments are conducted on the bike sharing dataset, which exhibits smaller log determinant values and therefore is more difficult to optimize. 
The results are summarized in Table \ref{tab:ablation}.
\begin{table}[h]
\centering
\begin{tabular}{l|cc|c|c}
Experiment  &min &pinv &train MSE &test MSE\\
\hline
1&-&-&8919 (4095) &9010 (4056)\\
2 &+ &- &1826 (660) &2092 (662)\\
3 &- &+ &2199 (1972) &2450 (1955)\\
4 (Algorithm \ref{alg:anmin})  &+ &+ &1457 (319)    &1714 (385)\\
\hline 
\end{tabular}
\vskip -2mm
\caption{Ablation experiments. Average MSEs (with std) obtained on the bike-sharing dataset by the ANMIN Algorithm \ref{alg:anmin} with different parts removed.}
\label{tab:ablation}
\end{table}

In Experiment 1,  the ANMIN algorithm is stopped the first time when the log determinant is less than $-10,000$ and the final loss is reported. 

In Experiment 2, at each iteration when the log determinant is less than $-10,000$, the matrix $\bA$ is re-initialized with random values. The model with minimum train loss over all iterations is reported with its training and test loss values.

\begin{figure*}[ht]
\centering
\begin{tabular}{ccc}
\includegraphics[width=0.3\linewidth]{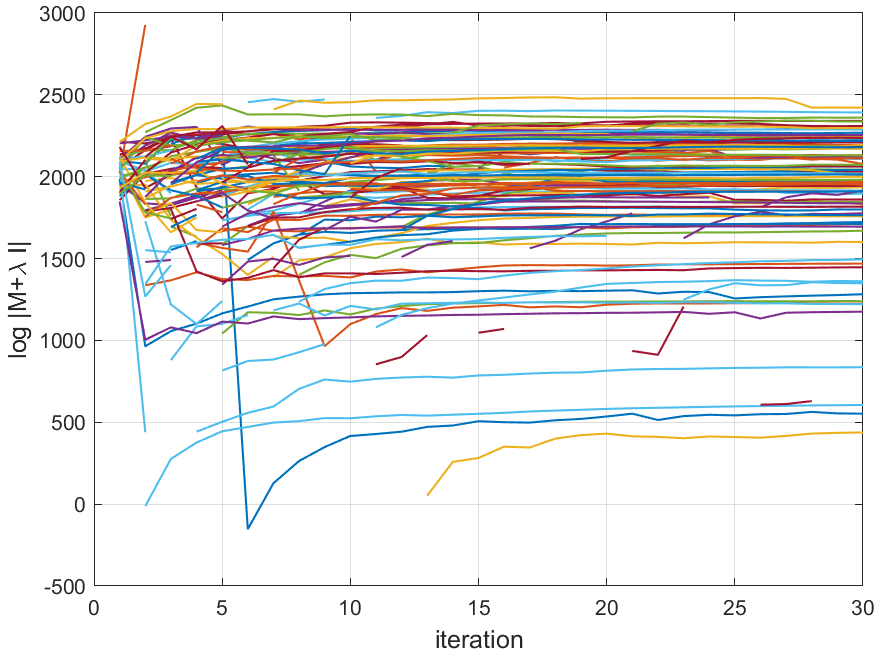}
&\includegraphics[width=0.3\linewidth]{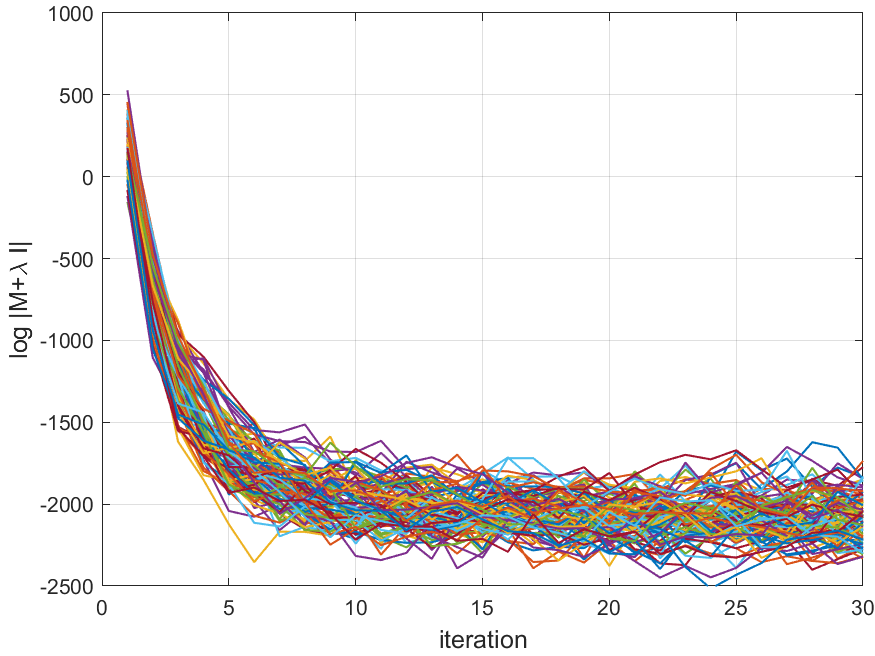}
&\includegraphics[width=0.3\linewidth]{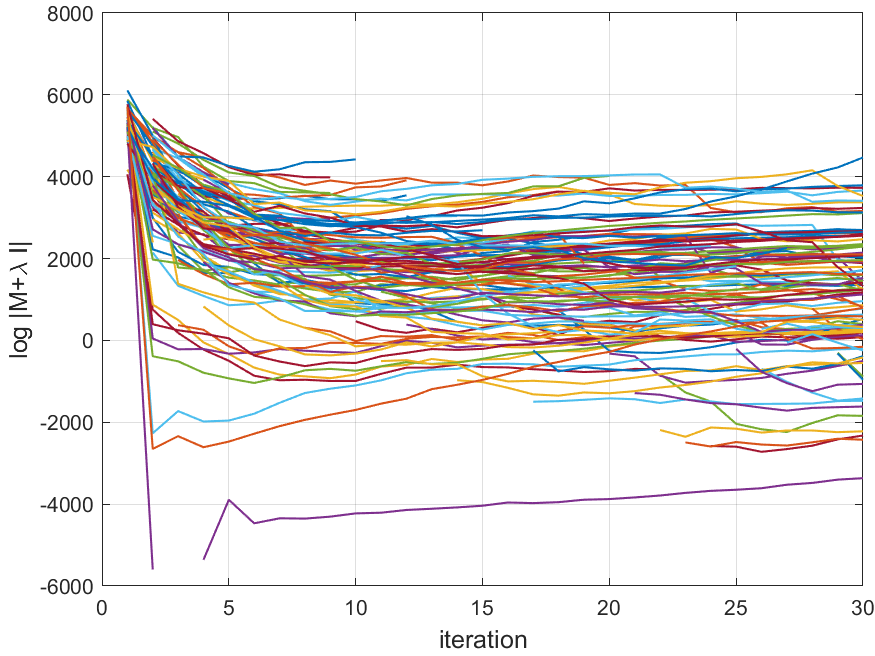}\vspace{-2mm}\\
SDF decoder  &abalone &bike-sharing 
\end{tabular}
\begin{tabular}{cc}
\includegraphics[width=0.3\linewidth]{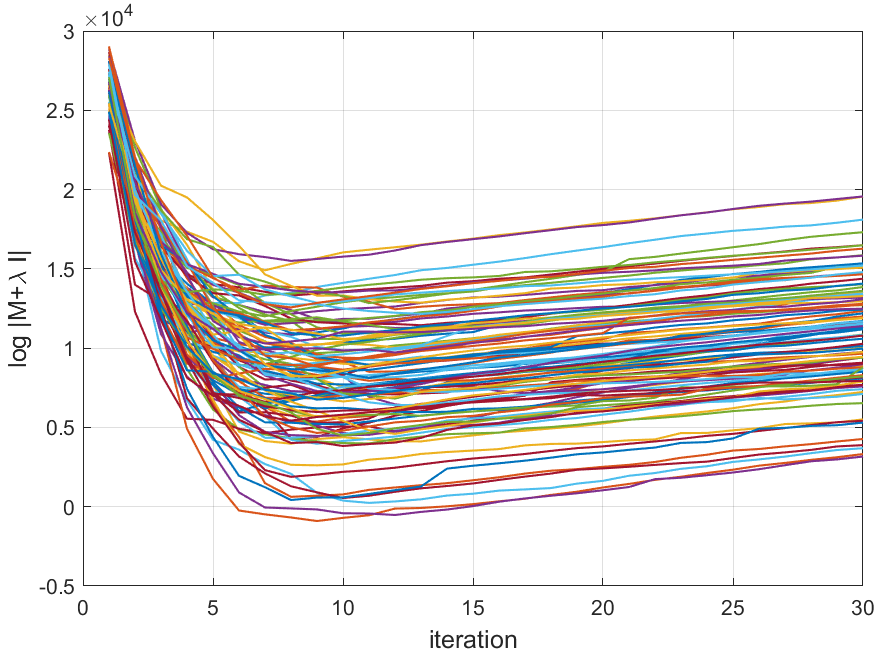}
&\includegraphics[width=0.3\linewidth]{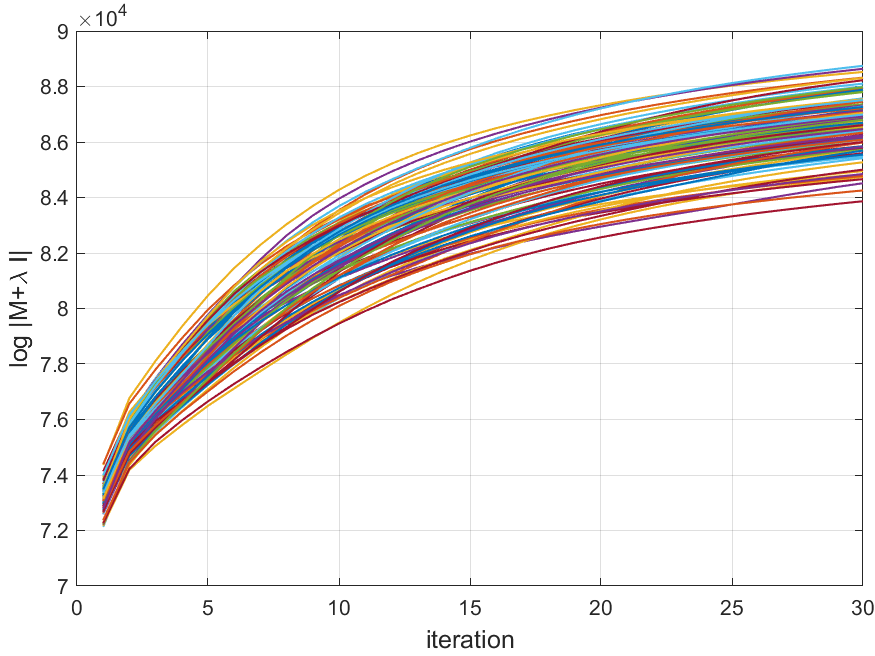}\\
Year pred MSD &DAE\\
\end{tabular}
\vspace{-2mm}
\caption{Plot of the $\ln |\bM+\lambda \bI_{(d+1)h}|$ vs iteration number for the 100 ANMIN (Algorithm \ref{alg:anmin}) runs for NN with ReLU activation.
} \label{fig:logdet}
\vspace{-2mm}
\end{figure*}

In Experiment 3, the algorithm is run as described in Alg. \ref{alg:anmin}, except that the final model is reported instead of the model with minimum loss over the iterations.

Experiment 4 is exactly as described in Algorithm \ref{alg:anmin}.

From Table \ref{tab:ablation} one could see that both the minimum over the iterations and the use of a pseudoinverse instead of a random restart help in obtaining a small final loss.

\begin{figure*}[ht]
\centering
\begin{tabular}{ccc}
\includegraphics[width=0.3\linewidth]{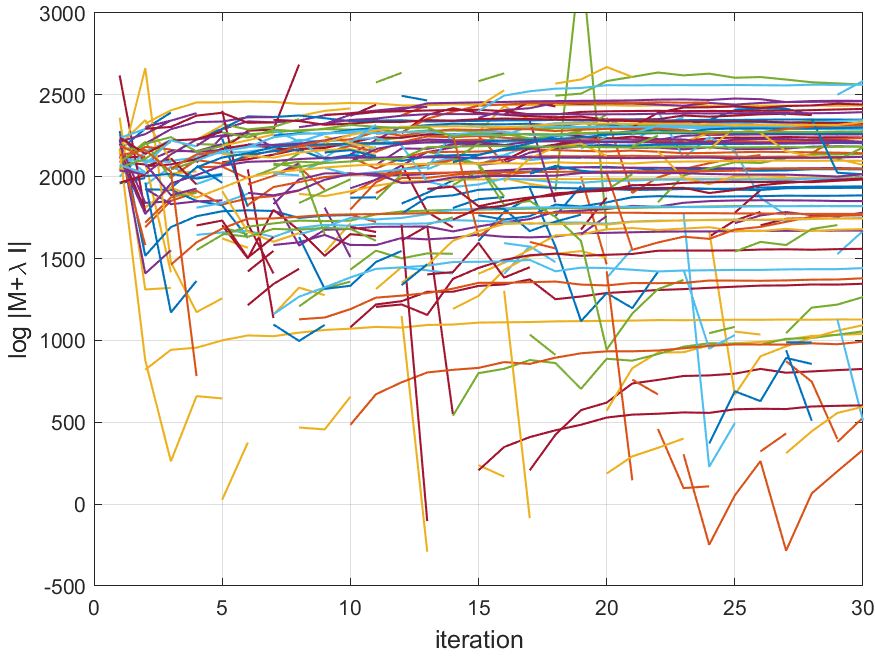}
&\includegraphics[width=0.3\linewidth]{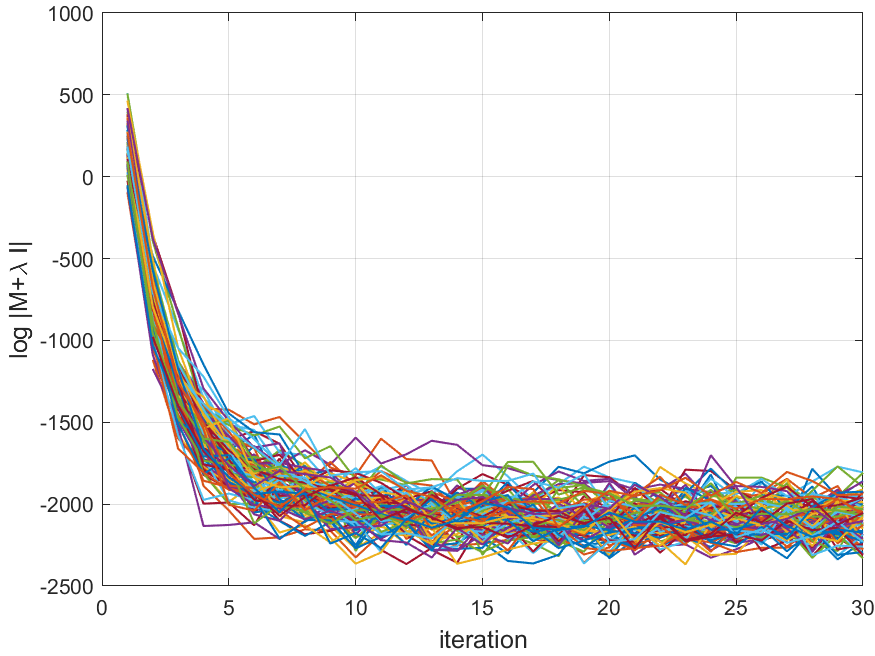}
&\includegraphics[width=0.3\linewidth]{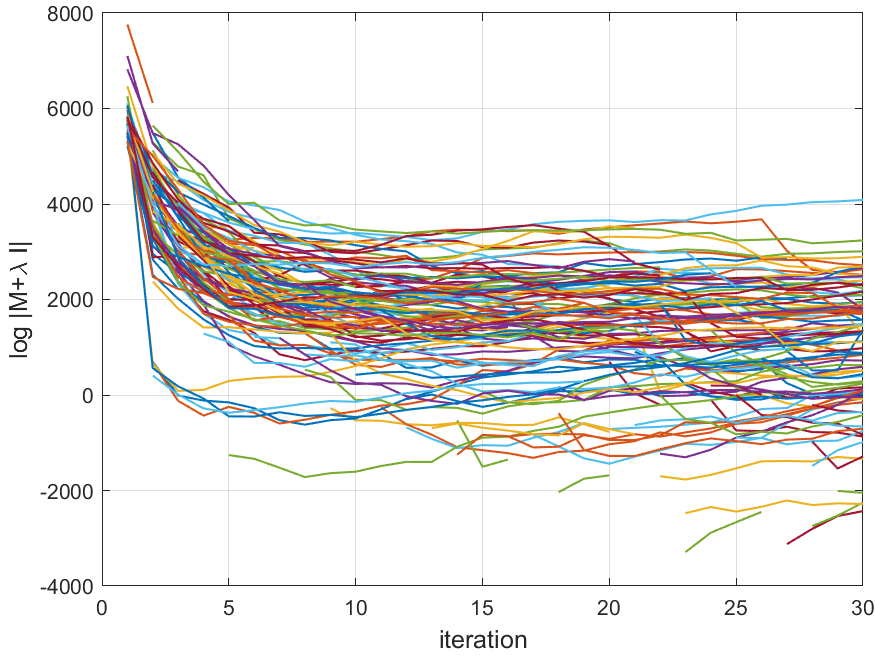}\vspace{-2mm}\\
SDF decoder  &abalone &bike-sharing 
\end{tabular}
\begin{tabular}{cc}
\includegraphics[width=0.3\linewidth]{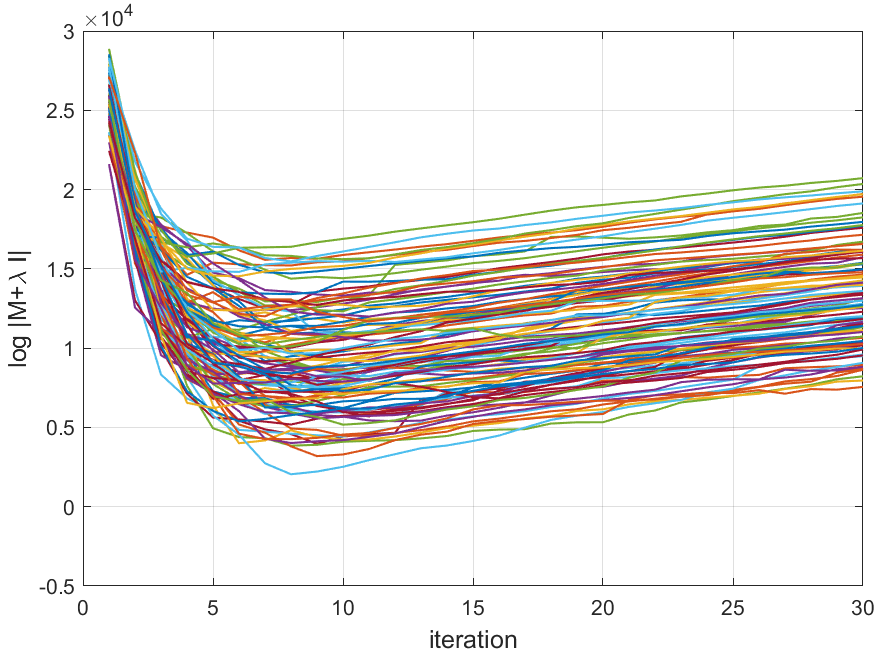}
&\includegraphics[width=0.3\linewidth]{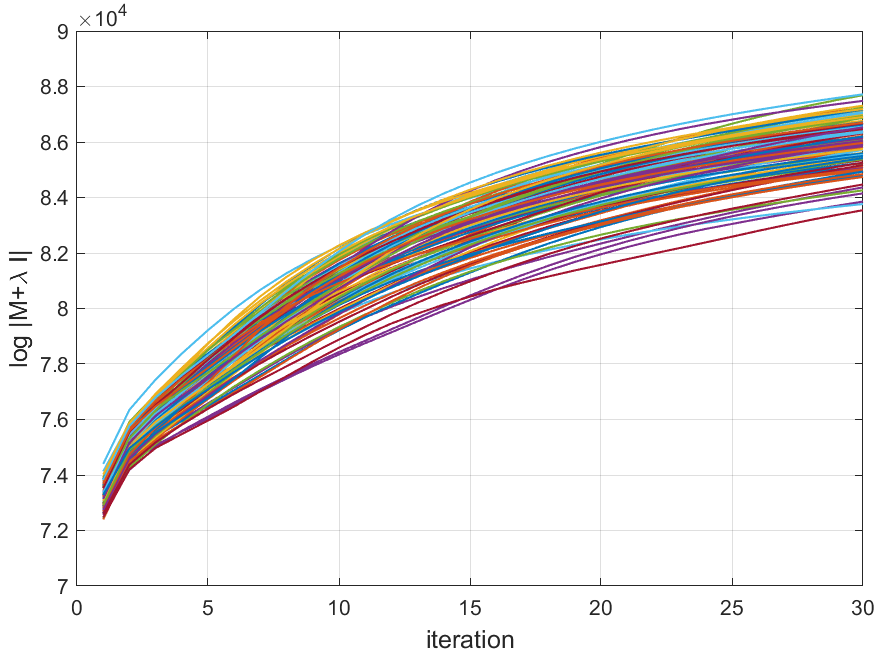}\\
Year pred MSD &DAE\\
\end{tabular}
\vspace{-2mm}
\caption{Plot of the $\ln |\bM+\lambda \bI_{(d+1)h}|$ vs iteration number for the 100 ANMIN (Algorithm \ref{alg:anmin}) runs for NN with leaky ReLU activation.
} \label{fig:logdetlk}
\vspace{-2mm}
\end{figure*}
In Figure \ref{fig:loss_bike} are shown the MSE training loss values for the SDF decoder and bike-sharing datasets obtained at each iteration (left) and the minimum MSE obtained so far (right). 
One could see that the MSE loss at each iteration sometimes jumps (usually when the log determinant is small or $-\infty$). 
However, the jumps are much smaller than those of the LBFGS method. 
This is because after each update of the matrix $\bA$, which can result in jumps, the matrix $\bB$ is also updated to minimize the MSE.

\subsection{Diagnosing the ANMIN Algorithm}

As opposed to gradient based algorithms, the ANMIN method offers some  capabilities to diagnose the energy landscape during the optimization. 
For that, the condition number of $\bM+\lambda \bI_{(d+1)h}$ or the log determinant $\ln |\bM+\lambda \bI_{(d+1)h}|$ can be used to see how degenerate is the local optimum that the ANMIN reaches when fitting the hidden nodes $\bA$. 
In Figure \ref{fig:logdet} are shown the log determinant values $\ln |\bM+\lambda \bI_{(d+1)h}|$ encountered during optimization for the 100 ANMIN runs on the random splits.
Some curves are broken because the log determinant is $-\infty$ at some iterations. 
One could see that for the one-output datasets, the log determinant has a tendency to decrease, while for the multiple output DAE problem, the log determinant increases with the iterations. 
For the low dimensional datasets (abalone and bike-sharing) the log determinant reaches a very small value, which makes the matrix $\bM+\lambda \bI_{(d+1)h}$ close to singular and the solution of Eq. \eqref{eq:A} less stable.

\section{Conclusion} \label{sec:conclusion}

This paper introduced a method for training a two layer NN with ReLU or leaky ReLU-like activation using the square loss by alternatively fitting the coefficients of each layer  while keeping the other layer and the activation pattern of the neurons fixed. 

Experiments indicate that the method can obtain in some cases (when the input dimension is small) strong minima of the loss function that go beyond the capabilities of those obtained by gradient descent optimization. 
It can do so with a relatively small number of iterations (10-30), which means that it can work with large datasets that don't fit in the computer's memory.

This work is experimental in nature, in that we introduce an algorithm and we observe that it works well experimentally, but we provide no proofs or theoretical guarantees that it will always work under certain assumptions.

This work could usher  the possibility of using shallow NNs for many computation sensitive applications, by providing better optimization capabilities for shallow NNs that go beyond those of gradient descent algorithms. 

It is know that tree ensembles,  such as boosted trees and random forests, can be mapped to two-layer neural networks \cite{sethi1991decision,welbl2014casting}. 
In this respect, this work could potentially enable better algorithms for training such deep ensembles by loss minimization, which are better at minimizing the loss and more computationally efficient.

However, the method has certain limitations since it involves the inversion of a matrix of size $(d+1)h\times (d+1)h$. 
This could limit the applicability of the method when the input dimension and the number of neurons are both large (e.g. $dh>30,000$). 
This drawback could be addressed by sparse neuron connections, where not all neurons are connected to all inputs, or by large scale implementations that use multiple GPUs.

In the future we plan to apply the ANMIN method to shape analysis using NNs and to large scale object detection applications.

\funding{This research received no external funding.}




\dataavailability{The following publicly available datasets have been used in experiments: Abalone, Bike-sharing and Year Prediction MSD from the UC Irvine Machine Learning Repository \cite{dua2019}, the Weizmann horse dataset \cite{borenstein2004combining}, and the Berkeley dataset \cite{MartinFTM01}.
The DAE and SDF decoder data can be found at: \url{https://github.com/barbua/ANMIN/tree/main/data}.
} 


\conflictsofinterest{The authors declare no conflict of interest.} 



\abbreviations{The following abbreviations are used in this manuscript:\\

\noindent 
\begin{tabular}{@{}ll}
ANMIN &Analytic Minimization Algorithm\\
BFGS &Broyden–Fletcher–Goldfarb–Shanno algorithm\\
CNN &Convolutional neural network\\
DAE &Denoising auto-encoder\\
LBFGS &Limited memory BFGS algorithm\\
MSE &Mean Square Error\\
NN &Neural network\\
OLS &Ordinary Least Squares\\
SGD &Stochastic gradient descent\\
ReLU &Rectified linear unit\\
\end{tabular}
}

\end{paracol}

\reftitle{References}


\externalbibliography{yes}
\bibliography{refs}
\end{document}